\documentclass{article}
\usepackage[accepted]{icml2023}
\usepackage{times}

\usepackage{amsmath,amsfonts,bm}

\def\eqref#1{equation~\ref{#1}}

\def\1{\bm{1}}

\DeclareMathAlphabet{\mathsfit}{\encodingdefault}{\sfdefault}{m}{sl}
\SetMathAlphabet{\mathsfit}{bold}{\encodingdefault}{\sfdefault}{bx}{n}

\newcommand{\E}{\mathbb{E}}

\DeclareMathOperator*{\argmin}{arg\,min}

\DeclareMathOperator{\X}{\mathcal{X}}
\DeclareMathOperator{\Y}{\mathcal{Y}}

\newcommand\indep{\protect\mathpalette{\protect\independenT}{\perp}}
\def\independenT#1#2{\mathrel{\rlap{$#1#2$}\mkern2mu{#1#2}}}

\usepackage[utf8]{inputenc} 
\usepackage[T1]{fontenc}    
\usepackage{hyperref}       
\usepackage[capitalize,noabbrev]{cleveref}
\usepackage{url}            
\usepackage{booktabs}       
\usepackage{amsfonts}       
\usepackage{nicefrac}       
\usepackage{microtype}      

\usepackage{nccmath}
\usepackage{bm}
\usepackage{menukeys}
\usepackage{multirow}
\usepackage{graphicx}
\usepackage{mathrsfs}
\usepackage{paralist}
\usepackage{subfig}
\usepackage{amsthm}
\usepackage{enumitem}
\usepackage{xspace}
\usepackage[normalem]{ulem}

\usepackage{amsmath}
\usepackage[ruled, algo2e]{algorithm2e}
\theoremstyle{plain}

\theoremstyle{definition}

\theoremstyle{remark}

\usepackage{wasysym}
\usepackage{siunitx}

\definecolor{fig_red}{RGB}{255, 122, 122}
\definecolor{fig_blue}{RGB}{143, 170, 220}
\definecolor{pearDark}{HTML}{2980B9}

\SetKwInput{KwInput}{Input}
\SetKwInput{KwOutput}{Output}
\SetKwFor{For}{for (}{)}{}

\usepackage{bbm}
\usepackage{float}

\usepackage{caption}
\usepackage{booktabs}

\usepackage{rotating}
\usepackage{dsfont}
\usepackage{tabularx}

\usepackage{thm-restate}

\usepackage[
  separate-uncertainty = true,
  multi-part-units = repeat
]{siunitx}

\usepackage{wrapfig}
\usepackage{pdfcomment}

\hypersetup{
  colorlinks,
  linkcolor={red!50!black},
  citecolor={blue!50!black},
  urlcolor={blue!80!black}
  }

\makeatletter
\renewcommand\paragraph{\@startsection{paragraph}{4}{\z@}%
{0.6ex \@plus.2ex \@minus.2ex}%
{-1em}%
{\normalfont\normalsize\bfseries}}
\makeatother

\usepackage[title]{appendix}

\usepackage{verbatim}

\usepackage{amssymb}
\usepackage{pifont}
\newcommand{\cmark}{\ding{51}}%
\newcommand{\xmark}{\ding{55}}
\newcommand{\x}{\mathbf{x}}
\newcommand{\method}{InstaAug\xspace}
\makeatletter
\newcommand*{\rom}[1]{\expandafter\@slowromancap\romannumeral #1@}

\def\genbox#1#2#3#4#5#6{
    \leavevmode\raise#4bp\hbox to#5bp{\vrule height#5bp depth0bp width0bp
    \pdfliteral{q .5 w \csname #2COLOR\endcsname\space RG
                       \csname #3PDF\endcsname{#5}{#6} S Q
             \ifx1#1 q \csname #2COLOR\endcsname\space rg 
                       \csname #3PDF\endcsname{#5}{#6} f Q\fi}\hss}}

\def\trianbox   #1#2{\genbox{#1}{#2}  {trian}    {0}   {5}    {2.5}}

\makeatother

\icmltitlerunning{Learning Instance--Specific Augmentations by Capturing Local Invariances}

\renewcommand\footnotemark{}

\begin{document}
\twocolumn[
\icmltitle{Learning Instance--Specific Augmentations by Capturing Local Invariances}
\icmlsetsymbol{equal}{*}

\begin{icmlauthorlist}
\icmlauthor{Ning Miao}{oxf}
\icmlauthor{Tom Rainforth}{oxf}
\icmlauthor{Emile Mathieu}{oxf}
\icmlauthor{Yann Dubois}{stan}
\icmlauthor{Yee Whye Teh}{oxf}
\icmlauthor{Adam Foster}{micro}
\icmlauthor{Hyunjik Kim}{deep}
\end{icmlauthorlist}

\icmlaffiliation{oxf}{Dept. of Statistics, University of Oxford}
\icmlaffiliation{stan}{Stanford University}
\icmlaffiliation{micro}{Microsoft Research}
\icmlaffiliation{deep}{DeepMind, UK}


\icmlcorrespondingauthor{Ning Miao}{ning.miao@stats.ox.ac.uk}

\icmlkeywords{Machine Learning, ICML}

\vskip 0.3in
]
\renewcommand{\thefootnote}{\arabic{footnote}}
\setcounter{footnote}{0}

\printAffiliationsAndNotice{}

\begin{abstract}

We introduce \emph{\method}, a method for automatically learning input-specific augmentations from data.
Previous methods for learning augmentations have typically assumed independence between the original input and the transformation applied to that input.
This can be highly restrictive, as the invariances we hope our augmentation will capture are themselves often highly input dependent.
\method instead introduces a learnable \emph{invariance module} that maps from inputs to tailored transformation parameters, allowing local invariances to be captured.
This can be simultaneously trained alongside the downstream model in a fully end-to-end manner, or separately learned for a pre-trained model.
We empirically demonstrate that \method learns meaningful input-dependent augmentations for a wide range of transformation classes, which in turn provides better performance on both supervised and self-supervised tasks. 
\end{abstract}

\section{Introduction}
\label{sec:intro}
Data augmentation is an important tool in deep learning~\citep{shorten2019survey}.
It allows one to incorporate inductive biases and invariances into models \citep{chen2019invariance,lyle2020benefits}, providing an effective regularization technique that aids generalization~\citep{goodfellow2016deep}. 
It has proved particularly successful for computer vision tasks, forming an essential component of many modern supervised  \citep{krizhevsky2012imagenet,perez2017effectiveness, mikolajczyk2018data, cubuk2020randaugment} and self-supervised  \citep{bachman2019learning,chen2020simple,tian2020makes,foster2021improving} approaches.

\begin{figure}[t]
 \centering
 \subfloat[Color jittering (red line indicates class boundary)]{
 \label{fig:idea_color}
 \centering
 \includegraphics[width=0.45\textwidth,trim={0 0.25cm 0 2cm},clip]{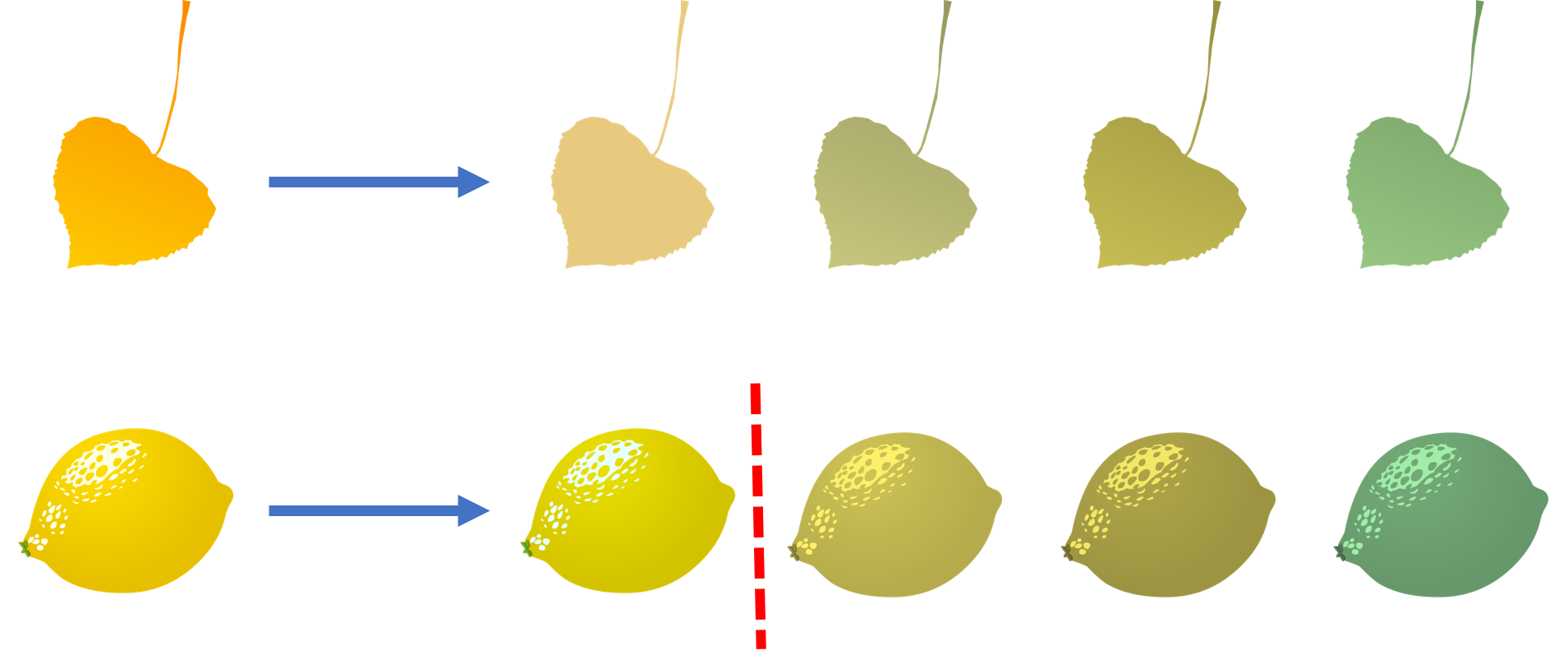}
 }
 
 \subfloat[Cropping]{
 \label{fig:idea_crop}
 \centering
 \includegraphics[width=0.45\textwidth]{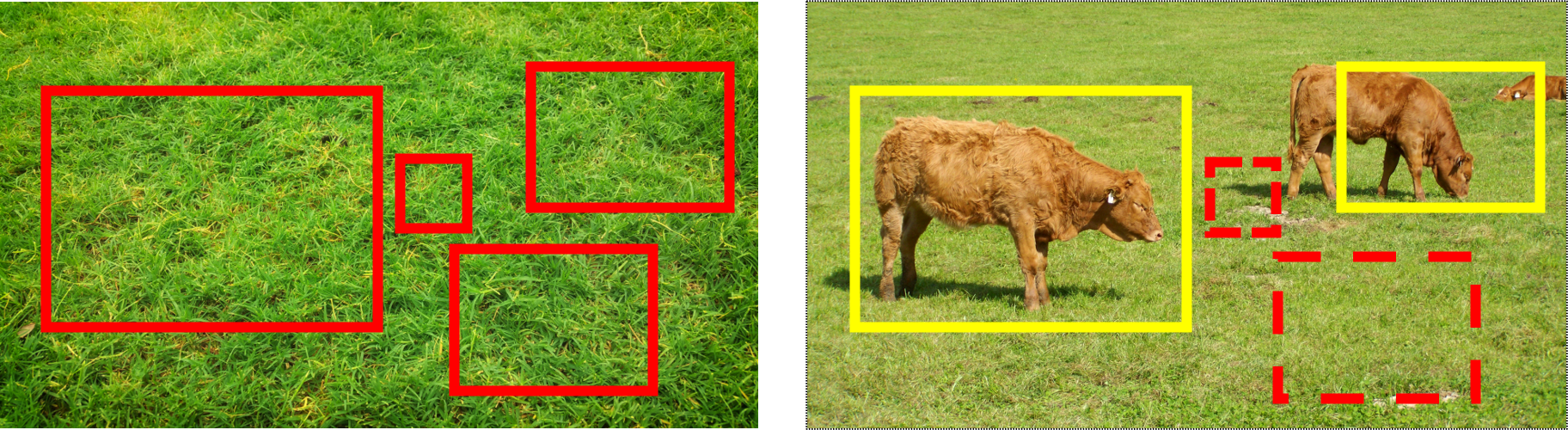}
 }
 \caption{
Different inputs require different augmentations. In (a), a leaf is invariant to color change from yellow to green, but the same transformation changes a lemon to a lime. 
In (b), the same effect is shown for cropping. Solid rectangles represent the patches that preserve the labels of the original images~([left] grass, [right] cattle), while dashed rectangles represent patches with different labels.
\vspace{-6pt}
}
 \label{fig:idea}
\end{figure}

Algorithmically, data augmentations apply a \emph{random transformation} $\tau: \X \rightarrow \X$, $\tau\sim p(\tau)$, to each input data point $\x \in \X$, before feeding this \emph{augmented} data into the downstream model.
These transformations are resampled each time the data point is used (e.g.~at each training epoch), effectively populating the training set with additional samples.
Augmentation is also sometimes used at test time by ensembling predictions from multiple transformations of the input.
A particular augmentation is defined by the choice of the \emph{transformation distribution} $p(\tau)$, whose construction forms the key design choice.
Good transformation distributions induce substantial and wide-ranging changes to the input, while preserving the information needed for prediction.

To try and ensure a good augmentation scheme, previous work has looked to \emph{learn} this transformation distribution from data
\citep{cubuk2018autoaugment,lim2019fast,benton2020learning}.
However, existing approaches typically assume independence between the input $\x$ and the transformation distribution $p(\tau)$.
As such, they are only able to learn \emph{global invariances}, severely limiting their flexibility and potential impact.
For example, when using color jittering, changing the color of a leaf from yellow to green would preserve its label, but the same transformation would change a lemon to a lime~(see \cref{fig:idea_color}).
This transformation cannot, therefore, be usefully applied as a global augmentation, 
even though it is a useful invariance for the \emph{specific input instance} of a leaf.
Similar examples regularly occur for other transformations, such as cropping (see~\cref{fig:idea_crop}).

Another line of recent work~\citep{zhou2020meta, cheung2022adaaug} has instead looked to utilize instance-aware augmentations by defining a small predefined set of allowable transformations, then introducing a policy that assigns probabilities (and magnitudes) to elements of this set as a function of the input.
While these approaches allow some of the shortfalls of global augmentations to be overcome, they do not have the flexibility to learn fine-grained transformation distributions, or uncover underlying invariances.

To address these shortfalls, we introduce \method, a new approach to learn \emph{instance-specific} augmentations by capturing \emph{local} invariances that are specific to the region of the provided input.
\method is based on using a transformation distribution of the form $p(\tau; \phi(\x))$, where $\phi$ is a deep neural network that maps inputs to transformation distribution parameters.
We refer to $\phi$ as an \emph{invariance module}. 
It can be trained simultaneously with the downstream model in a fully end-to-end manner, or individually with a fixed pre-trained model.
Both cases only require access to training data and optimize a single objective function that minimizes the training error while maintaining transformation diversity.
As such, \emph{\method} allows one to directly learn powerful and general augmentations, without requiring access to additional data or annotations.

We evaluate \method in both supervised and self-supervised settings, focusing on image classification and contrastive learning respectively.
Our experimental results show that \method is able to uncover meaningful invariances that are consistent with human cognition, and improve model performance for various tasks compared with baseline models. 
While we primarily focus on the case where the invariance module is trained alongside the downstream model (to allow augmentation during training), we find that \method can also provide substantial performance gains when used to learn test-time augmentations for large pre-trained models.
Accompanying code is provided at {\small\url{https://github.com/NingMiao/InstaAug}}.


\vspace{-3pt}
\section{Background}
\label{sec:background}

Data augmentation methods operate as a wrapper algorithm around some downstream model, $f$, randomly transforming the inputs $\x\in\X$ before they are passed to the model.
The outputs of the augmented model are given by $f(\tau(\x))$, where $\tau : \X \mapsto \X$ represents the transformation, sampled from some transformation distribution $p(\tau)$.
The aim of this augmentation is to instill inductive biases into the learned model, leading to improved generalization by capturing invariances of the problem.
It can be used both during training to provide additional synthetic training data, and/or at test-time, where ensembling the predictions from multiple transformations can provide a useful regularization that often improves performance~\citep{shanmugam2021better}.

Some approaches look to learn aspects of the augmentation~\citep{cubuk2018autoaugment,cubuk2020randaugment,lim2019fast,ho2019population,hataya2020faster,li2020dada,zheng2022deep}.
These approaches can be viewed as learning parameters of $p(\tau)$, helping to automate its construction and tuning.  
Of particular relevance, Augerino~\citep{benton2020learning} provides a mechanism for learning augmentations using a simple end-to-end training scheme, where the parameters of the downstream model and transformation distribution are learned simultaneously using the (empirical) risk minimization 
\begin{align}\label{eq:augerino}
    \hspace{-3pt}\min\nolimits_{f,\theta}  \mathbb{E}_{\x,y \sim p_{\text{data}}} \left[\mathbb{E}_{\tau \sim p_{\theta}(\tau)} \left[\mathcal{L}(f(\tau(\x)),y) \right] \right]\!+\!\lambda \mathbb{R}(\theta),
\end{align}
where $\mathcal{L}$ is a loss function and $\lambda \mathbb{R}(\theta)$ is a regularization term that encourages large transformations.

All of these approaches can be thought of as \emph{global} augmentation schemes, in that transformations are sampled independently to the input.
For an unrestricted, universal, class of transformations, this assumption can be justified through the noise outsourcing lemma~\citep{kallenberg1997foundations}: any conditional distribution $Y|X=\x$ can be expressed as a deterministic function $g: \X \times\, \mathbb{R} \to \Y$ of the input and some independent noise $\varepsilon \sim \mathcal{N}(0,I)$.
Thus, using reparameterization, the dependency on $\x$ can, in principle, be entirely dealt with by the transformation itself.
However, in practice, the transformation class must be restricted to provide the desired inductive biases, meaning this result no longer holds and so the independence assumption can cause severe restrictions.
For example, sampling color jitterings independently to the input is equivalent to the unrealistic assumption that the labels of all images $\x$ are invariant to the same group of changes~(\textit{cf.} \cref{fig:idea_color}).

\section{InstaAug: Capturing Local Invariances} \label{sec:method}

In order to remedy the problems of global augmentations, we propose \method.
\method learns an \emph{input dependent} distribution $p(\tau; \phi(\x))$ of information-preserving transformations that actively makes use of the input $\x$ via the \emph{invariance module} $\phi$, as opposed to learning a global transformation distribution $p_{\theta}(\tau)$.
This generalizes the hypothesis class of transformation distributions, and significantly increases the flexibility and expressivity of the resulting augmentation, without undermining our ability to carefully control the inductive biases that are imparted.
It can also informally be viewed as a mechanism for learning invariances that are local to the specific input.

We argue that a good augmentation strategy needs to fulfill two properties.
First, the transformations should preserve the information in $\x$ that is necessary for the task at hand.
For example, transformations must preserve information about the label for supervised tasks.
Second, the set of transformations needs to have sufficient `diversity' to effectively augment the data; we quantify this as the entropy of the transformation distribution $p(\tau; \phi(\x))$.
In addition to their intuitive nature, in~\Cref{sec:app:theory} we provide theoretical analysis that shows these requirements naturally originate from a decomposition of the generalization error between the true risk and augmented empirical risk of $f$.
For simplicity, we describe \method for the specific case where $f$ is a classifier in the remainder of this section.

\subsection{Model structure}
\begin{figure}
  \vspace{5pt}
 \centering
 \includegraphics[width=0.45\textwidth]{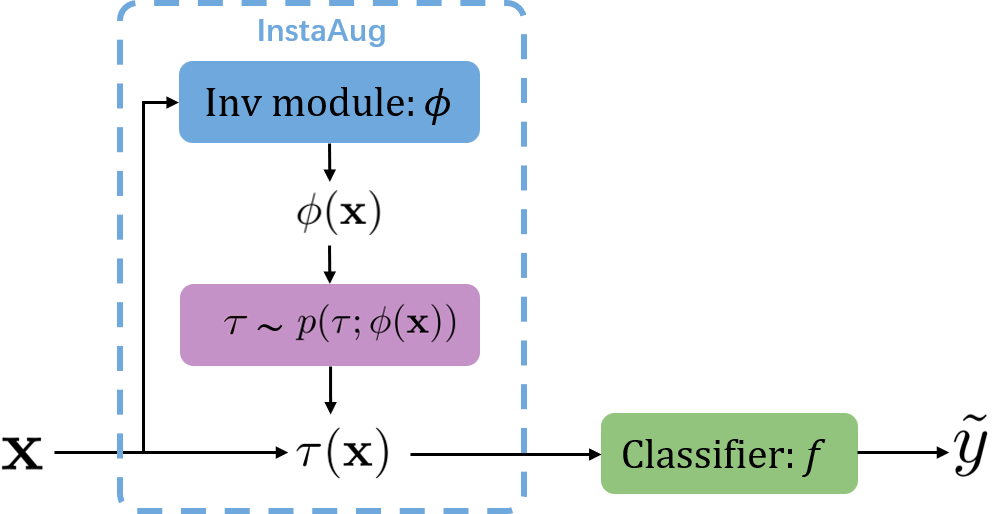}
 \caption{Summary of \method. 
 }
 \label{fig:model}
 \vspace{-5pt}
\end{figure}
\method is based around using a simple plug-in invariance module, $\phi$, between the input $\x$ and the classifier $f$, as shown in~\cref{fig:model}. 
We assume a parametric family of distributions $p(\tau; \cdot)$ over some transformation space, then use $\phi$, which is a trainable neural network, to predict its parameters for a given input.
During training, we \emph{sample} a transformation $\tau \sim p(\tau;\phi(\x))$, which is applied to $\x$ to generate an augmented sample $\tau(\x)$, before feeding this into the classifier $f$.

\subsection{Training}
\label{sec:method_learn}

\begin{figure}
 \vspace{-16pt}
 \centering
 \subfloat[Global augmentation]{
 \label{fig:loss_1}
 \centering
 \includegraphics[width=0.23\textwidth]{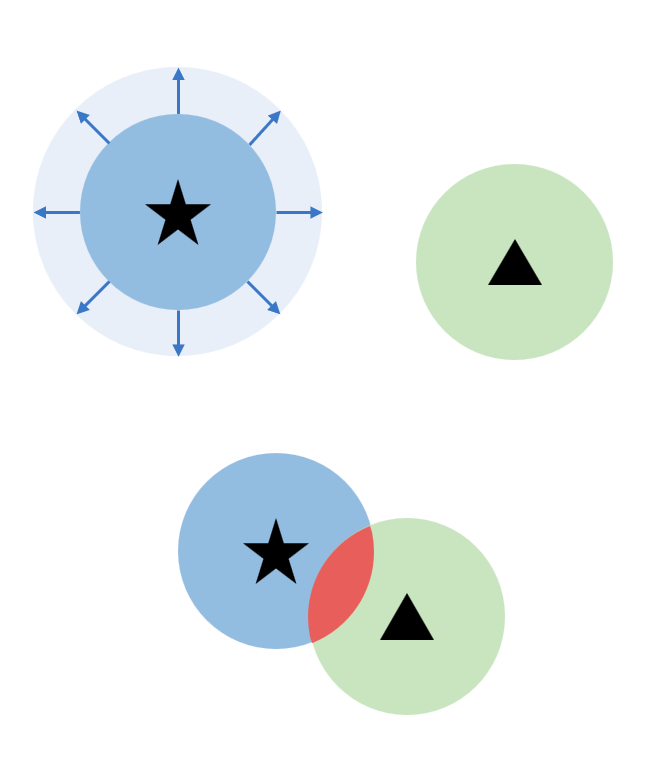}
 }
 \subfloat[\method]{
 \label{fig:loss_2}
 \centering
 \includegraphics[width=0.23\textwidth]{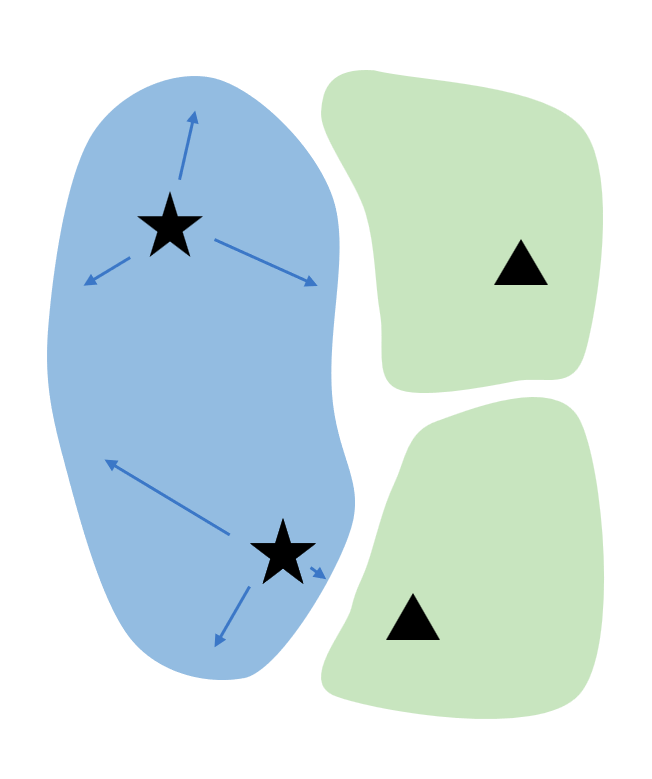}
 }
 \caption{\method learns more diverse augmentations that also preserve labels compared to global augmentations. $\bigstar$ and \trianbox{1}{cblack} are samples from two different classes. Blue and green shades represent label-preserving augmentations for each class.
 In (a), the upper $\bigstar$ would benefit from being further augmented, but some of the augmented samples for the lower $\bigstar$ are already over-augmented and indistinguishable from another class~(see the red intersection). \method solves this problem by learning a different augmentation for each instance, as shown in (b).}
 \label{fig:loss}
 \vspace{-5pt}
\end{figure}

Good augmentations should induce substantial changes to the input $\x$ while preserving all necessary information about the task at hand, thereby capturing the maximum possible invariance.
\cref{fig:loss_1} illustrates the tension between these two objectives experienced by global augmentation schemes.
Wider-ranging transformations are generally beneficial for generalization, but `excessive' transformations will generate samples that will be incorrectly classified.
In \cref{fig:loss_1} we see this in the red area, where the augmentations for a pair of data points have started to overlap, creating ambiguity and inevitably misclassifications.
Using instance-specific augmentations (\cref{fig:loss_2}) allows for a better trade-off of these needs.
However, to achieve this we need our objective to encourage diversity in augmentations, not just low training error.
It should also let the level of diversity vary between inputs, as some points will be able to support larger transformations than others.

Based on these needs, training is done by simultaneously minimizing a conventional expected loss with respect to both $\phi$ and $f$ (or just $\phi$ if $f$ is a fixed pre-trained classifier as per~\Cref{sec:exp:fixed}), while regularizing the average entropy of the transformations,~$\mathbb{E}_{\x \sim p_{\text{data}}}~[\mathbb{H}[ p(\tau; \phi{(\x)})]]$.
The core motivation for this setup is that minimizing the expected loss will naturally encourage the information needed for prediction to be preserved, but the regularization on entropy is needed to enforce diversity. Further motivation is provided by the theoretical analysis of~\Cref{sec:app:theory}.

By appropriately parameterizing $p(\tau; \phi{(\x)})$ (see \cref{sec:method_parameterization}), we can write down its entropy in closed form.
We can then formulate the problem as minimizing the following w.r.t.~$f$ and $\phi$:
\vspace{3pt}
\begin{align}
\mathbb{E}_{\x,y \sim p_{\text{data}}, \tau \sim p(\tau; \phi(\x))} \left[\mathcal{L}(f(\tau(\x)),y) 
\!-\!\mathbf{\lambda} 
\mathbb{H}[p(\tau; \phi(\x))]
\right],
\label{eqn:main}
\end{align}\vspace{3pt}where $\mathcal{L}$ is the loss of the downstream task, for which we will generally use the cross-entropy. 
Unlike in the Augerino objective of~\cref{eq:augerino}, $\lambda$ here is an automatically-tuned weight of the entropy term that enables precise control over transformation diversity. 
Specifically, we initialize $\lambda$ with a small positive value, then increase it when the average entropy drops below a lower bound $\text{H}_{\min}$ and decrease it when it exceeds an upper bound $\text{H}_{\max})$ during training. 
Here $\text{H}_{\min}$ and $\text{H}_{\max}$ are hyperparameters, through which we can directly control the diversity level of learned transformations during the whole training process. As described in \cref{apptab:H_ablation}, they can easily be tuned.

This dynamic $\lambda$ is necessary because the requirement for $\lambda$ is different at different stages of training.
In the beginning, when the classifier is weak, we need a small $\lambda$ to avoid the transformations becoming overly diverse, which results in different classes overlapping with each other. As the classifier gets more powerful during training, larger $\lambda$ is needed to compete with the cross-entropy term $\mathcal{L}$. 

Using this approach, the invariance module and downstream model can be trained simultaneously using end-to-end gradient descent, utilizing the reparameterization trick to deal with the stochasticity of $\tau$ when possible~\citep{kingma2013auto}, and the REINFORCE estimator~\citep{williams1992simple} otherwise.
The approach can also be extended to regression or self-supervised learning by substituting the loss function $\mathcal{L}$ (see \Cref{sec:app:method}).

\subsection{Parameterization of augmentations}
\label{sec:method_parameterization}
The parameterization method is another critical factor in the quality of learned invariances. A good parameterization should be flexible enough to reflect the complexity of real data while not creating obstacles to gradient-based learning.
Here we focus on parameterizing transformations that are frequently used in computer vision, though our framework can easily be extended to other domains.
Due to the varied characteristics of different image transformations, we design two different parameterization methods for $p(\tau; \phi(\x))$.

\begin{figure}
 \centering
 \includegraphics[width=0.45\textwidth]{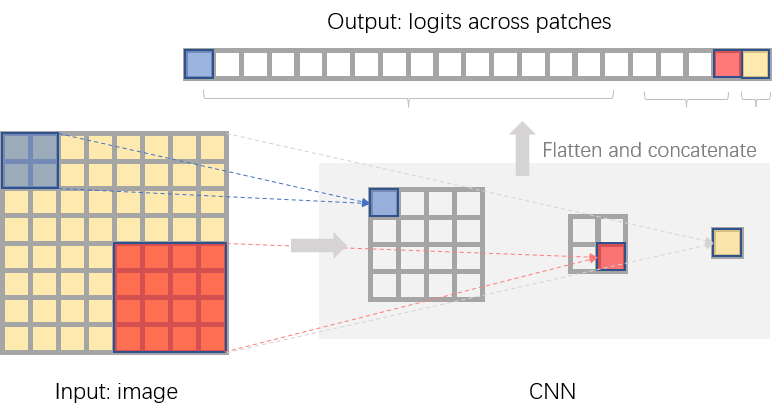}
 \caption{Location-related parameterization of crops by a CNN. The shaded area (bottom right) shows a simplified 3-layer CNN, and squares represent units at different convolutional layers.
 Each unit defines a patch in the input image (shown in the same color) through its receptive field.
 The activation value of the unit then gives the corresponding unnormalized log probability for that patch.
 }
 \label{fig:parameterization}
 \vspace{-7pt}
\end{figure}

\textbf{Uniform parameterization.}~~
For simpler transformations, such as rotation and color jittering, we find that a uniform distribution is enough for parameterizing $p(\tau; \phi(\x))$, such that
$\phi(\x)$ returns a pair $(\theta_{\min}, \theta_{\max})$ representing extrema of the possible transformations.
For example, for rotations, these represent the maximum and minimum rotation angles, such that $\tau(\x) = R(\theta)\x$, where $\theta \sim \mathcal{U}(\theta_{\min}, \theta_{\max})$ and $R(\theta)$ is the rotation operator.
To compose multiple transformations,
we simply sample them independently, such that $p(\tau_1, \dots, \tau_K; \phi(\x)) = \prod_{k=1}^K p(\tau_k; \phi_k(\x))$.
This provides a similar parameterization to~\citep{benton2020learning}, but where $(\theta_{\min}, \theta_{\max})$ now critically varies with the input $\x$ and there is no symmetry assumption on this range.

\textbf{Location-related parameterization}~~ 
Using this uniform parameterization is unfortunately not appropriate for more complex transformations like cropping.
Firstly, the distribution of crop centers may be multi-modal, since important information may exist in different parts of an image. Secondly, the desired crop size and center are often highly correlated so cannot be sampled independently. 
Finally, we encountered significant practical training issues when using the uniform parameterization for cropping, with $\phi$ becoming trapped in local optima with little transformation diversity.

We, therefore, propose an alternative location-related parameterization (LRP) for cropping, which is based on defining a large set of representative crops, then constructing $\phi$ to map from inputs to a vector of probabilities over this set.
As shown in~\cref{fig:parameterization}, this is achieved using a CNN where each hidden unit corresponds to a possible crop defined by its receptive field.
In order to select crops with different sizes, units from different layers are utilized, with those of earlier/latter layers representing smaller/larger crops.
This parameterization proved more effective than simply outputting the probabilities from a conventional network, due to the greater parameter sharing between related crops.
We note that it can also be directly extended to other transformations, such as masking, local blurring, pixel-wise perturbation, and local color jittering.

\vspace{-2pt}
\section{Related Work}
\label{sec:related}

\textbf{Hard-coded invariance.}~ Many recent works have looked to hard-code global invariance in neural networks. For example, various architectures have been designed to be invariant to translation~\citep{chaman2021truly,zhang2019making}, rotation~\citep{worrall2017harmonic,zhou2017oriented, marcos2017rotation}, scaling~\citep{worrall2019deep, sosnovik2019scale} or other group actions~\citep{cohen2016group,xu2021group}. 
Unfortunately, they require the set of invariant transformations to be closed under composition, leaving out many practical transformations that do not form a group.

\textbf{Learning augmentations.}~ 
There have been numerous prior works that automatically learn \emph{global} augmentations and invariance from data. 
As discussed in~\Cref{sec:background}, Augerino~\citep{benton2020learning} is perhaps the most closely linked such approach to \method as it also relies on end-to-end training (see~\Cref{sec:ap:augerino} for further discussion on its similarities and differences to \method).
AutoAugment~\citep{cubuk2018autoaugment} instead uses reinforcement learning to find augmentation strategies that increase accuracy on a separate validation set. Various follow-up works have improved its efficiency and/or performance~\citep{lim2019fast,ho2019population, tang2019adatransform,hataya2020faster,li2020dada,cubuk2020randaugment,zheng2022deep}. 

\begin{figure}
 \centering
 ~
 \subfloat[Augerino]{
 \label{fig:rotation_augerino}
 \centering
 \includegraphics[width=0.18\textwidth]{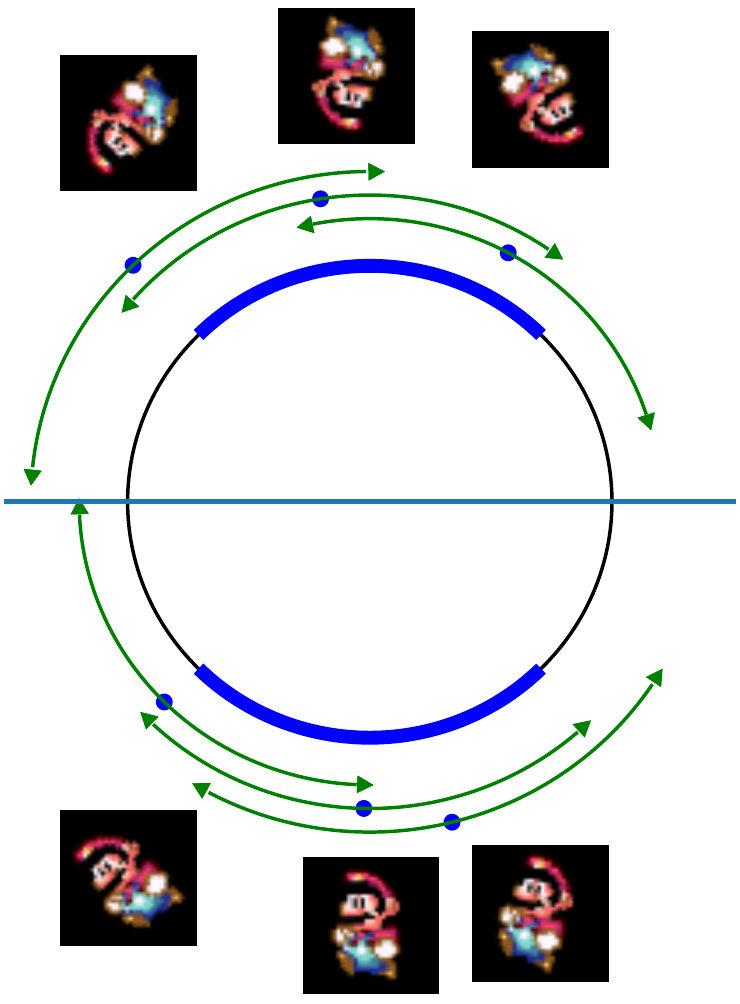}
 }
 \hspace{0.02\textwidth}
 \subfloat[\method \ (Ours)]{
 \label{fig:rotation_ours}
 \centering
 \includegraphics[width=0.18\textwidth]{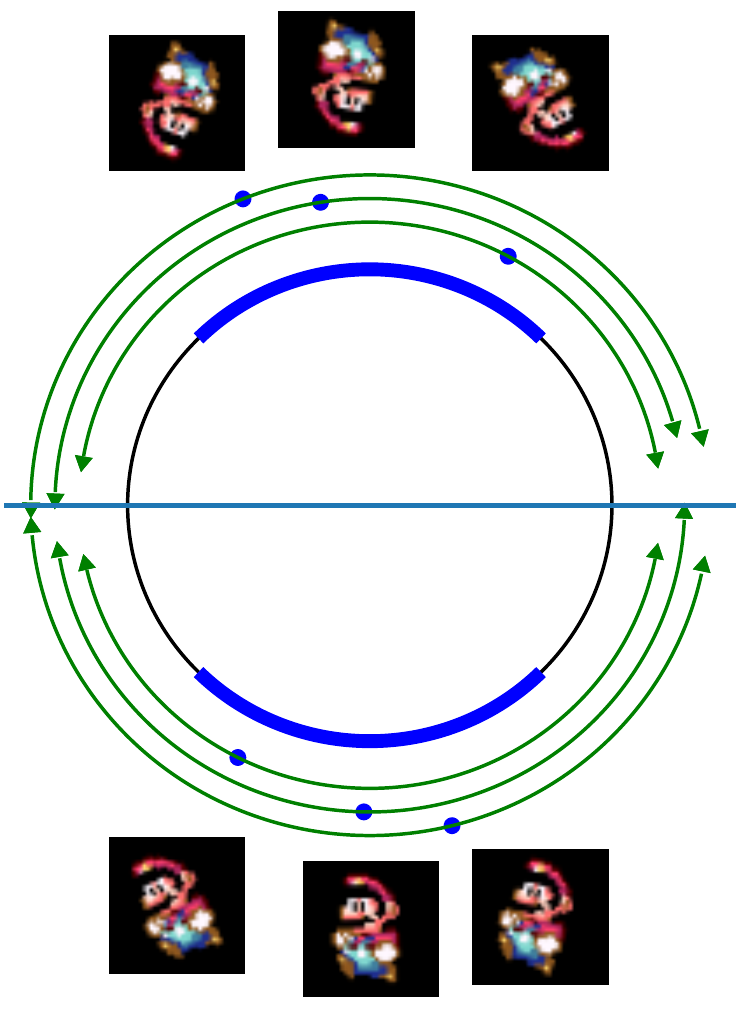}
 }
 \caption{
 Learned invariances for the Mario and Iggy dataset. 
 The blue arcs show the training data range, while the green arcs show example learned transformation distributions.
 }
 \label{fig:rotation}
 \vspace{-7pt}
\end{figure}

\textbf{Augmentation policies.}~
A couple of recent works have further looked to learn augmentation \emph{policies} that allow a degree of dependency on the input or class label, namely AdaAug~\citep{cheung2022adaaug} and MetaAugment~\citep{ZhouLXCHSL21}.
These policies assign probabilities and magnitudes to a fixed finite list of possible transformation operations.
Though they can depend on the input, they only make discrete choices and cannot learn a fine-grained transformation distribution in the way that \method does; they are thus not suitable for capturing local invariances.
For example, using \method with cropping learns a joint distribution over patch positions and sizes, whereas these methods only learn a probability for whether to apply cropping or not, and a scalar magnitude to use if it is applied. 
Further, both AdaAug and MetaAugment require a separate validation set, only consider augmentation during training (so cannot be used for pre-trained classifiers), and cannot be applied in unsupervised settings.
AdaAug also has the additional restriction that its policy is based on a linear mapping from the penultimate layer of the classification model, so its input dependence is inherently limited.
Meanwhile, although MetaAugment learns a sample-level policy network, in practice it averages this policy among training samples to form a global policy applied to all samples.

\textbf{Other related work.}
The spatial transformer~\citep{jaderberg2015spatial} aims to learn instance-specific transformations, but only applies a single transformation to each input rather than a distribution of transformations, making it distinct from data augmentation.
\citet{luo2020learn} and \citet{kim2020learning} both also learn instance-specific augmentations. However, the latter consider only test-time augmentation, while the former introduces an approach that is highly specialized to test recognition and cannot be applied in the more general settings we consider.
\citet{tamkin2020viewmaker} and~\citet{chen2021robust} both utilize adversarial augmentations to increase robustness.
\citet{zhou2020meta} learn symmetries shared across several datasets through a meta-learning scheme.
Mixup methods~\citep{zhang2018mixup, yun2019cutmix, rame2021mixmo} can also be thought of as a specific type of data augmentation.
Some of them~\citep{kim2020puzzle,kim2020co,park2022saliency}, allow for input dependence through gradient-based saliency~\citep{simonyan2013deep}. However, they use a fixed augmentation strategy rather than learning a transformation distribution.

\section{Supervised Learning Experiments} \label{sec:experiment}

\subsection{Rotated 2D images}
We first consider a simple synthetic dataset proposed in \citet{benton2020learning}.
The dataset contains four categories, (1) upright Mario; (2) upside-down Mario; (3) upright Iggy; and (4) upside-down Iggy. Each of the four base images is randomly rotated in the interval of $[-\pi/4, \pi/4]$ to form the training dataset. The task is to predict the correct character (Mario vs Iggy) and the orientation (up vs down).
We assess whether \method is able to learn the `best' rotation range for each sample---i.e. the maximum range that avoids `up' and `down' classes from overlapping.

\cref{fig:rotation} shows that \method effectively recovers the broadest range of rotations for each image while preserving labels, while Augerino only learns a subset of these ranges.
This can be most easily seen by the fact that the transformation distributions (shown in green) always extend to very close to the true class boundary for \method, but not for Augerino.
These gains are because
Augerino learns a \emph{single} global augmentation distribution shared across all images (note the shared transformation distribution arcs), which are inevitably limited for any given input.

\subsection{Cropping} \label{sec:exp_cropping}
We now move to more realistic images and to the most common and effective form of image augmentation: cropping.
We first evaluate the performance of jointly training \method and the classifier on Tiny-Imagenet (TinyIN, $64 \times 64$), as it inherits the image complexity of ImageNet whilst being within our computational budget.
TinyIN is a standard testbed for data augmentations.
Full experiment details are given in~\Cref{apd:exp_details_cropping}.

We benchmark \method alongside several augmentation baselines, including random crop, Augerino, and AdaAug.
For random crop, we use a uniform distribution on patch size and position, tuning the bounds on the former through a comprehensive hyperparameter sweep to ensure appropriate scaling.
We further compare to other prior works that have obtained competitive results on TinyIN~\citep{zhang2018mixup, yun2019cutmix,rame2021mixmo, kim2020puzzle, kim2020co, park2022saliency}, directly taking their reported results.

\setlength\intextsep{0pt}
\begin{table}
\vspace{7pt}
    \centering
    \small
    \caption{\label{tab:crop}
Test accuracy for Tiny-ImageNet with cropping augmentation.
The Instance column indicates whether the method is instance-specific or not.
Statistics are computed over 10 runs, except for MixUp methods, whose results are directly taken from their respective papers. 
We omit comparison to other global augmentation schemes, as these only learn the size ranges of the cropping, which is already covered by the hyperparameter tuning of Random crop.
}
\begin{tabular}{ lcc } 
 \toprule
 Method\ & Instance  & Accuracy (\%) \hspace{-4pt}\
 \\ \midrule
  No augmentation             &\xmark& $55.06${\tiny $\pm 0.10$} \\
  Random crop         &\xmark& $64.49${\tiny $\pm 0.12$}  \\
  \midrule
  MixUp              &\xmark& $63.74$\\
  CutMix             &\xmark& $65.09$ \\
  MixMo              &\xmark& $64.80$ \\
  Puzzle-Mix         &\cmark& $63.48$\\
  Co-Mixup           &\cmark& $64.15$\\
  Saliency grafting    &\cmark& $64.84$\\
  
  \midrule
  Augerino &\xmark& $55.02${\tiny $\pm 0.29$}  \\
  AdaAug             &\cmark& $64.03${\tiny $\pm 0.19$} \\
  \quad\quad -w/  LRP         &\cmark& $62.01${\tiny $\pm 0.23$}  \\
  \method (ours)              &\cmark & {$\textbf{66.02}${\tiny $\pm 0.18$}} \\%
  \quad\quad -w/o LRP &\cmark&   $55.39${\tiny $\pm 0.19$}\\
  \quad\quad -w/o input         &\xmark& $63.20${\tiny $\pm 0.12$} \\
  \quad\quad -class-specific         &\xmark/\cmark& $60.55${\tiny $\pm 0.50$} \\
  
 \bottomrule
\end{tabular}
\end{table}
In order to ablate the effects of input-dependency and our location-related parameterization (LRP, see \cref{fig:parameterization}) on \method, 
we additionally assess the performance of \textit{\method (without LRP)} by using same uniform parameterization as Augerino; 
\textit{\method (without input)} that uses the LRP and general InstaAug setup, but shares the transformation distribution across all inputs rather than learning an input-specific augmentation; and \textit{\method (class-specific)}, which takes training labels instead of images as inputs.
Test-time augmentation using $50$ transformation samples is deployed for all variants of \method, along with the Augerino, AdaAug, and random augmentation baselines (see \cref{sec:test-time}).
For \method (class-specific), this test-time augmentation is based on random cropping, as class information is not available at test-time and simply omitting test-time augmentation performed poorly.
Following prior works, we choose the PreActResNet-18 architecture~\citep{he2016identity} with $\text{width}=1$ as the classifier for all methods.

\cref{tab:crop} shows the (top-1) test accuracy for each method.
In agreement with prior works, we find that random cropping increases accuracy by 9.4\% over no augmentation, which is achieved where cropping scale $=[0.1, 1]$. \method outperforms random cropping and its own global version without input by 1.5\% and 2.8\% respectively, highlighting the effect of learning instance-specific augmentation.

\begin{figure}
 \centering
  \subfloat{
     \centering
     \begin{minipage}{0.25\textwidth}
       \center 
       {\small (A)~\method~(w/o input)}
      \end{minipage}}
    \hspace{0.015\textwidth}     
  \subfloat{
     \centering
     \begin{minipage}{0.18\textwidth}
       \center 
       {\small (B)~\method}
      \end{minipage}}   
      \hspace{0.03\textwidth} 
 \vspace{-18pt}
 
 \subfloat{
 \centering
 \includegraphics[width=0.0975\textwidth]{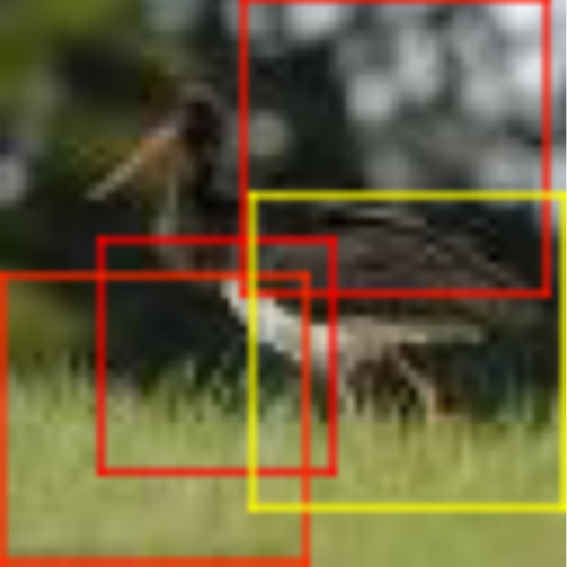}}
 \hspace{0.0025\textwidth}
 \subfloat{
 \centering
 \includegraphics[width=0.0975\textwidth]{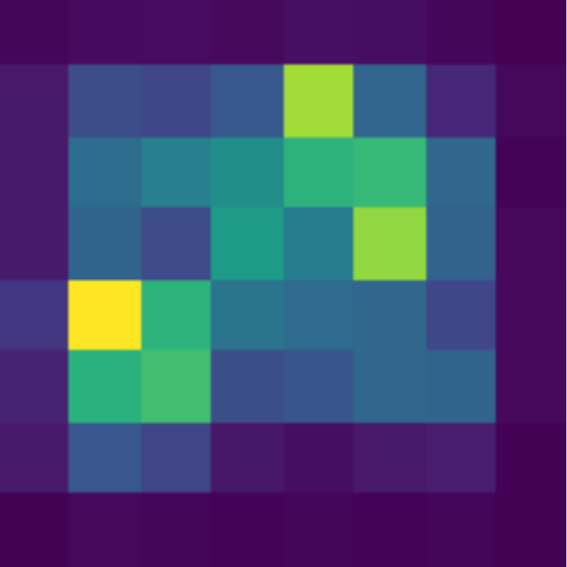}}
 \hspace{0.0025\textwidth}
 \subfloat{
 \centering
 \includegraphics[width=0.00975\textwidth]{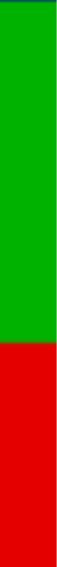}}
\hspace{0.005\textwidth}
 \subfloat{
 \centering
 \includegraphics[width=0.0975\textwidth]{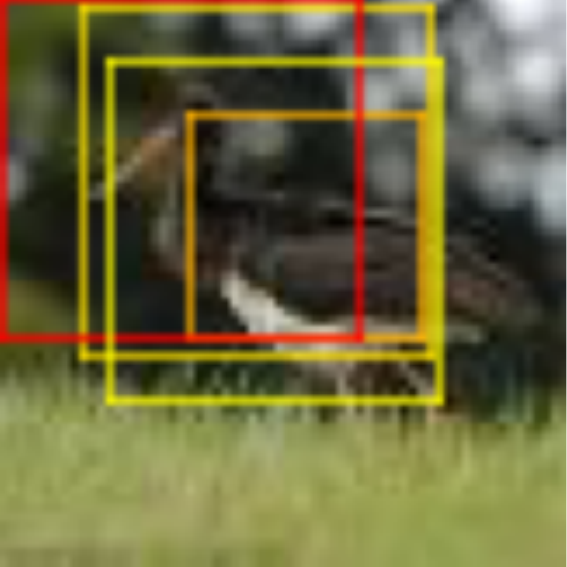}}
 \hspace{0.0025\textwidth}
 \subfloat{
 \centering
 \includegraphics[width=0.0975\textwidth]{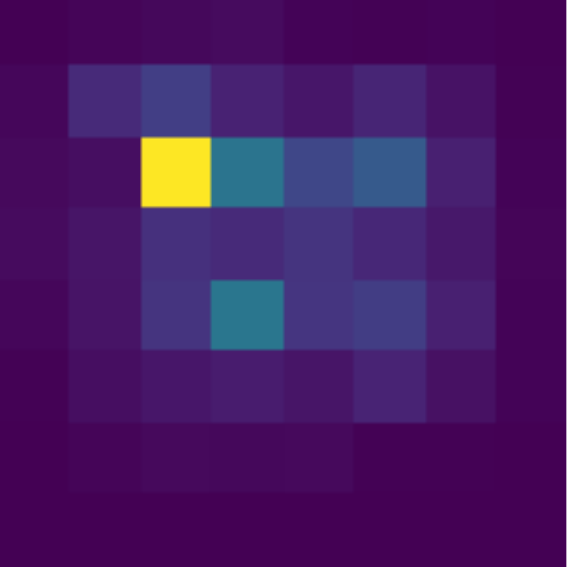}}
 \hspace{0.0025\textwidth}
 \subfloat{
 \centering
 \includegraphics[width=0.00975\textwidth]{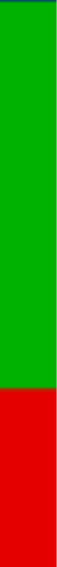}}
 \vspace{-6pt}
 
 \subfloat{
 \centering
 \includegraphics[width=0.0975\textwidth]{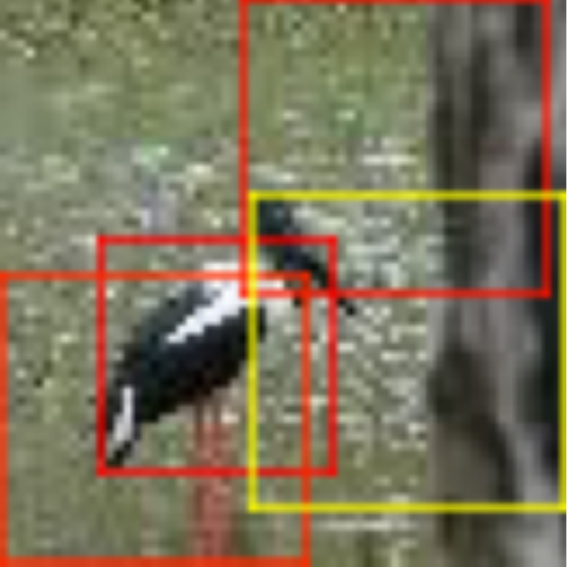}}
 \hspace{0.0025\textwidth}
 \subfloat{
 \centering
 \includegraphics[width=0.0975\textwidth]{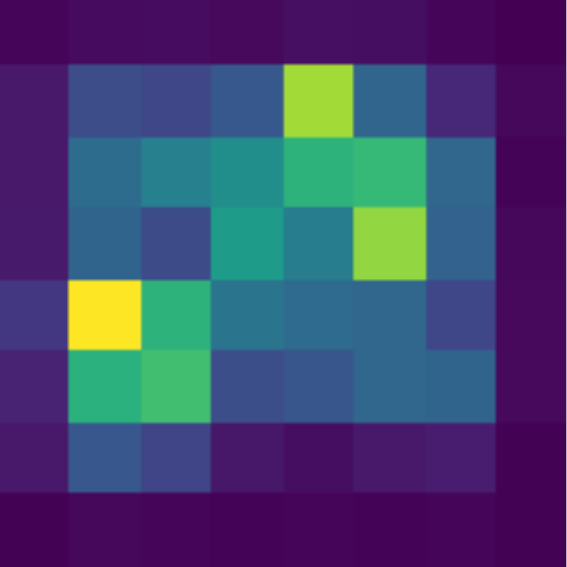}}
 \hspace{0.0025\textwidth}
 \subfloat{
 \centering
 \includegraphics[width=0.00975\textwidth]{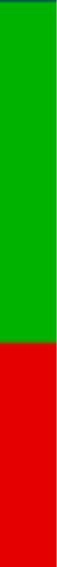}}
 \hspace{0.005\textwidth}
 \subfloat{
 \centering
 \includegraphics[width=0.0975\textwidth]{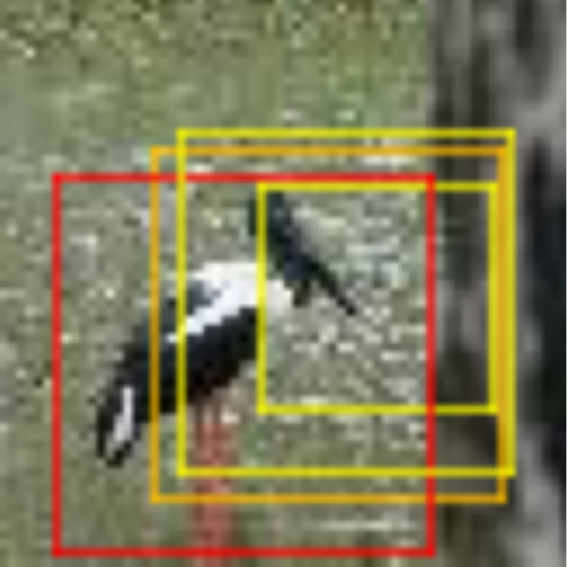}}
 \hspace{0.0025\textwidth}
 \subfloat{
 \centering
 \includegraphics[width=0.0975\textwidth]{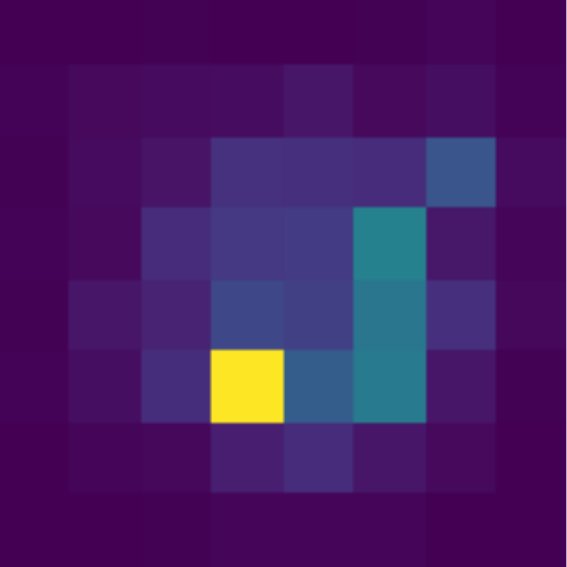}}
 \hspace{0.0025\textwidth}
 \subfloat{
 \centering
 \includegraphics[width=0.00975\textwidth]{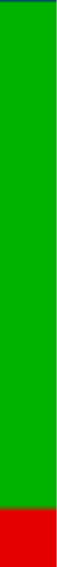}}
 \vspace{-6pt}
 
 \centering
 \setcounter{subfigure}{0}
 \subfloat[]{
 \centering
 \includegraphics[width=0.0975\textwidth]{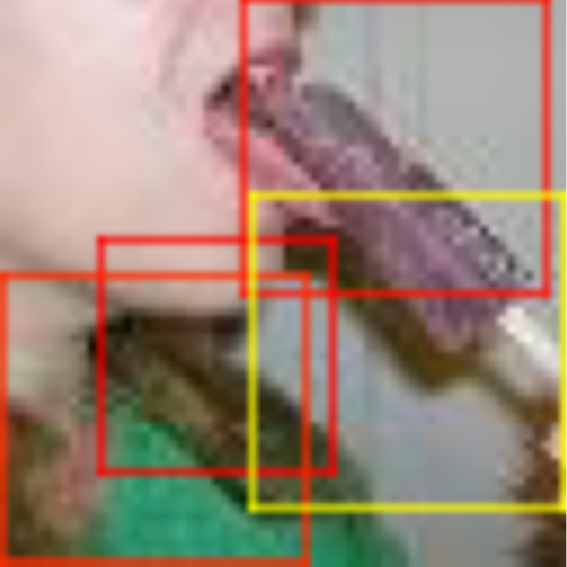}}
 \hspace{0.0025\textwidth}
 \subfloat[]{
 \centering
 \includegraphics[width=0.0975\textwidth]{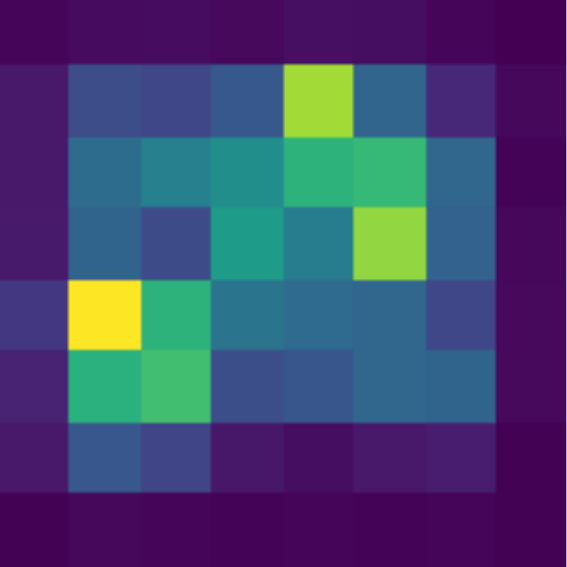}}
 \hspace{0.0025\textwidth}
 \subfloat[]{
 \centering
 \includegraphics[width=0.00975\textwidth]{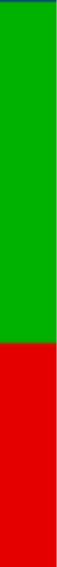}}
 \hspace{0.005\textwidth}
 \subfloat[]{
 \centering
 \includegraphics[width=0.0975\textwidth]{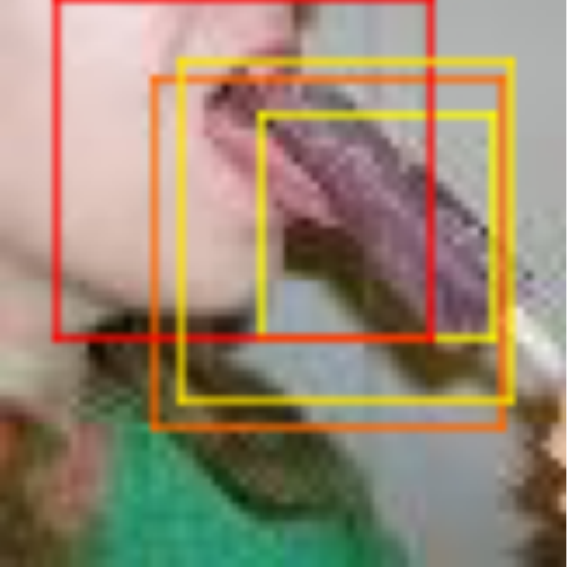}}
 \hspace{0.0025\textwidth}
 \subfloat[]{
 \centering
 \includegraphics[width=0.0975\textwidth]{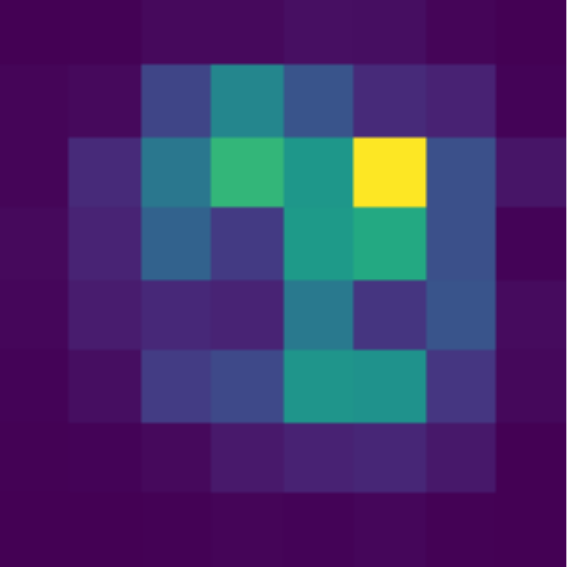}}
 \hspace{0.0025\textwidth}
 \subfloat[]{
 \centering
 \includegraphics[width=0.00975\textwidth]{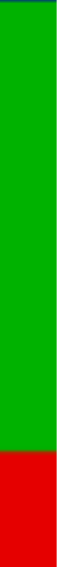}}
 
 \vspace{-3pt}
 \caption{\method~(B) learns more sensible crops compared to random and learned global~(A) augmentations. Columns (a, d) show examples of sampled crops, with red edges indicating higher probability.
 Columns (b, e) show density maps for the crop centers, with brighter color meaning higher probability.
 Columns (c, f) give the proportion of crops (red) above a particular size threshold, showing that \method produces fewer large crops.
 \vspace{-8pt}
 }
 \label{fig:crop_examples}
\end{figure}
Allowing only for class dependence actually produces even worse performance than just ignoring the input completely, presumably because of the inevitable resulting mismatch in the augmentations used in training and testing.
Methods with mean-field uniform parameterization~(including Augerino and \method without LRP)
performed extremely poorly, noticeably worse than just random cropping.
This is because they were found to become easily stuck at local minima with low cropping diversity, leading to similar performance as \textit{no augmentation}.
The original version of AdaAug achieves similar performance to Random crop, but is incapable of dealing with the large search space of LRP, leading to a small reduction in performance when this is added.
Note that the potentially unexpectedly good performance of the random cropping baseline compared to the other global baselines stems from the careful hyperparameter sweep used to tune its crop size, which proved more effective than these more direct training mechanisms. See~\Cref{app:sec:random} for more discussion.

\Cref{fig:crop_examples} shows example crops and learned transformation distributions for \method and a global augmentation scheme (\method without input).
We see that \method is able to learn a cropping scheme that focuses on the key aspect of the input image, while the baselines cannot.

\subsection{Applying \method to a fixed classifier}
\label{sec:exp:fixed}
\method can also be used to learn suitable augmentations for a fixed pre-trained classifier. This can most notably be useful as a means to learn test-time augmentations. As the invariance module is itself only a small network, it can be done relatively cheaply, even when the dataset and downstream model are very large.
We exploit this on the larger Imagenet dataset ($224 \times 224$)~\citep{deng2009imagenet}, again focusing on cropping augmentations and utilizing the LRP parameterization from~\cref{sec:method_parameterization}.

Training the invariance module in this setting is done in exactly the same way as elsewhere, using the training procedure of~\cref{sec:method_learn} with the normal training data.
The only thing that is changed is that $f$ is now fixed to a pre-trained classifier---specifically, the ResNet-50~\citep{he2016deep} from \citet{rw2019timm} (which did not use an invariance module during training)---rather than being simultaneously learned.
We are thus simply learning invariances, without affecting the training of $f$.

\begin{table}
    \centering
    \scriptsize
\caption{\label{tab:crop_imagenet}
\method boosts the test accuracy (\%) with test-time augmentation on Imagenet.
Invariance modules learned on ResNet-50 can also be directly applied to other models such as ResNet-18 and XCiT to improve generalization without fine-tuning. 
By contrast, global augmentation schemes are actually detrimental to test-time augmentation.
}
\begin{tabular}{ lcccc } 
 \toprule
 Method\ & \#Sample & ResNet50 & ResNet18 & XCiT\
 \\ \midrule
  No aug           &1&$80.43$& $69.73$ & $86.34$\\
   \midrule
  Random crop        &4&$78.45${\tiny $\pm 0.04$}& $66.13${\tiny $\pm 0.04$} & $82.05${\tiny $\pm 0.01$}\\
  AutoAug        &4&$77.84${\tiny $\pm 0.05$}& $59.50${\tiny $\pm 0.01$} & $81.40${\tiny $\pm 0.00$}\\
  FastAutoAug        &4&$77.87${\tiny $\pm 0.06$}& $61.43${\tiny $\pm 0.02$} & $81.42${\tiny $\pm 0.01$}\\
  \method            &4&$\textbf{80.92}${\tiny $\pm 0.04$}& $\textbf{70.59}${\tiny $\pm 0.05$} & $\textbf{86.43}${\tiny $\pm 0.04$}\\
  \midrule
  Random crop        &10&$79.60${\tiny $\pm 0.01$}& $67.87${\tiny $\pm 0.01$} & $82.84${\tiny $\pm 0.00$}\\
  AutoAug      &10&$79.20${\tiny $\pm 0.04$}& $63.96${\tiny $\pm 0.03$} & $82.43${\tiny $\pm 0.02$}\\
  FastAutoAug        &10&$79.28${\tiny $\pm 0.01$}& $65.65${\tiny $\pm 0.02$} & $82.45${\tiny $\pm 0.02$}\\
  \method            &10&$\textbf{81.18}${\tiny $\pm 0.02$}& $\textbf{70.96}${\tiny $\pm 0.03$} & $\textbf{86.47}${\tiny $\pm 0.02$}\\
  
 \bottomrule
\end{tabular}
\end{table}
In \cref{tab:crop_imagenet}, we show the effect of using the learned invariance module for test-time augmentation, finding that it is able to noticeably improve accuracy, unlike the baseline test-time augmentations of random cropping, AutoAugment~\citep{cubuk2018autoaugment}, and Fast AutoAugment~\citep{lim2019fast}.
Note that AdaAug cannot be used in this fixed-classifier setting.

In order to evaluate the generalization performance of our learned augmentation module, we further apply the augmentation trained on ResNet-50 to two different models \emph{with zero fine-tuning}: ResNet-18~\citep{he2016deep} and XCiT~\citep{ali2021xcit}. 
We find that the learned augmentation transfers very effectively to these different models, which implies that the local invariances \method learns to reflect the natural invariances of the underlying classification problem, rather than being specific to the model that was used to train the augmentation module.

\subsection{Color jittering on textures}\label{sec:exp_color}
Color jittering is another important type of data augmentation, which can help models generalize to different lighting conditions. 
We benchmark on the texture classification dataset RawFooT~\citep{bianco2017improving}.
RawFooT includes 68 different samples of raw food and each sample has an image taken under each of 46 lighting conditions (see \cref{fig:rawfoot_dataset_example} for examples), which makes it an ideal testbed to investigate methods' generalization ability between different lighting conditions.
We crop the original images to create the train set and test set. For each original image with a resolution of $800 \times 800$, we randomly sample 200 different $200 \times 200$ patches in the upper half as training images. The same procedure is taken on the lower half to produce test images, giving a train set and a test set for each different lighting condition. To evaluate the generalization ability to a broader range of lighting conditions, we evenly mix test images from all lighting conditions to form a general test set, while controlling the conditions during training. 

\begin{table}
    \centering
    \scriptsize
\caption{\method achieves higher general accuracy than baseline methods when trained on D45~(Daylight, 4500K).
}
\begin{tabular}{ lcc } 
 \toprule
 Method\ &Test aug? & Accuracy (\%)\
 \\ \midrule
  No aug             &\xmark& $72.87_{\pm 0.10}$ \\\midrule
  Random aug         &\xmark& $79.99_{\pm 0.13}$  \\
  Augerino        &\xmark& $78.97_{\pm 0.10}$ \\
  AdaAug        &\xmark& $75.27_{\pm 0.30}$ \\
  \method   &\xmark& $\textbf{81.11}_{\pm 0.20}$ \\\midrule
  Random aug &\cmark& $80.55_{\pm 0.16}$\\
  Augerino      &\cmark& $79.34_{\pm 0.14}$ \\
  AdaAug        &\cmark& $76.43_{\pm 0.15}$ \\
  \method   &\cmark&  $\textbf{81.35}_{\pm 0.19}$ \\%
 \bottomrule
\end{tabular}
\label{tab:color_overall_result}
\end{table}

\begin{table}[t]
    \centering
    \scriptsize
\caption{\method significantly outperforms baseline methods in general test accuracy~(\%) on different difficulty levels. Difficulty level is controlled by the number of randomly sampled lighting conditions seen.  Test-time augmentation is included for random and \method and we repeat each experiment for $10$ times.
}
\begin{tabular}{ lcccc } 
 \toprule
 \#Lighting conditions\hspace{-12pt} & 1 & 2 & 4 & 8\
 \\ \midrule
  No aug                  & ${68.5}_{\pm 2.6}$ & ${78.1}_{\pm 1.8}$ & ${84.8}_{\pm 0.7}$ & ${87.8}_{\pm 0.5}$ \\
  Random augmentation \hspace{-7pt}         & ${72.7}_{\pm 2.7}$ & ${80.8}_{\pm 1.3}$ & ${85.9}_{\pm 0.6}$ & ${87.3}_{\pm 0.3}$ \\
  \method   & ${\textbf{76.0}}_{\pm 2.5}$ & ${\textbf{83.6}}_{\pm 1.1}$ & ${\textbf{88.2}}_{\pm 0.5}$ & ${\textbf{89.6}}_{\pm 0.3}$ \\%
 \bottomrule
\end{tabular}

    \label{tab:color_difficulty_result}
\end{table}

We first train on a single lighting condition D45~(4500K, daylight) resembling natural light. \Cref{tab:color_overall_result} shows that \method outperforms all baselines with and without test-time augmentation.
We find that Augerino~(with relaxed symmetry restrictions on learned intervals) underperforms random augmentation because its parameters $\phi$ are often stuck near their initial values. We believe this is due to the conservative nature of using global augmentations (\textit{cf.} \Cref{fig:loss}), where even a small change in the parameters may largely increase the training loss, which prohibits wide-ranging augmentations. AdaAug does not perform well either, which might be a result of its inability to learn the interval length for the distribution of each transformation.

\begin{figure}
 \centering
 \includegraphics[width=0.45\textwidth]{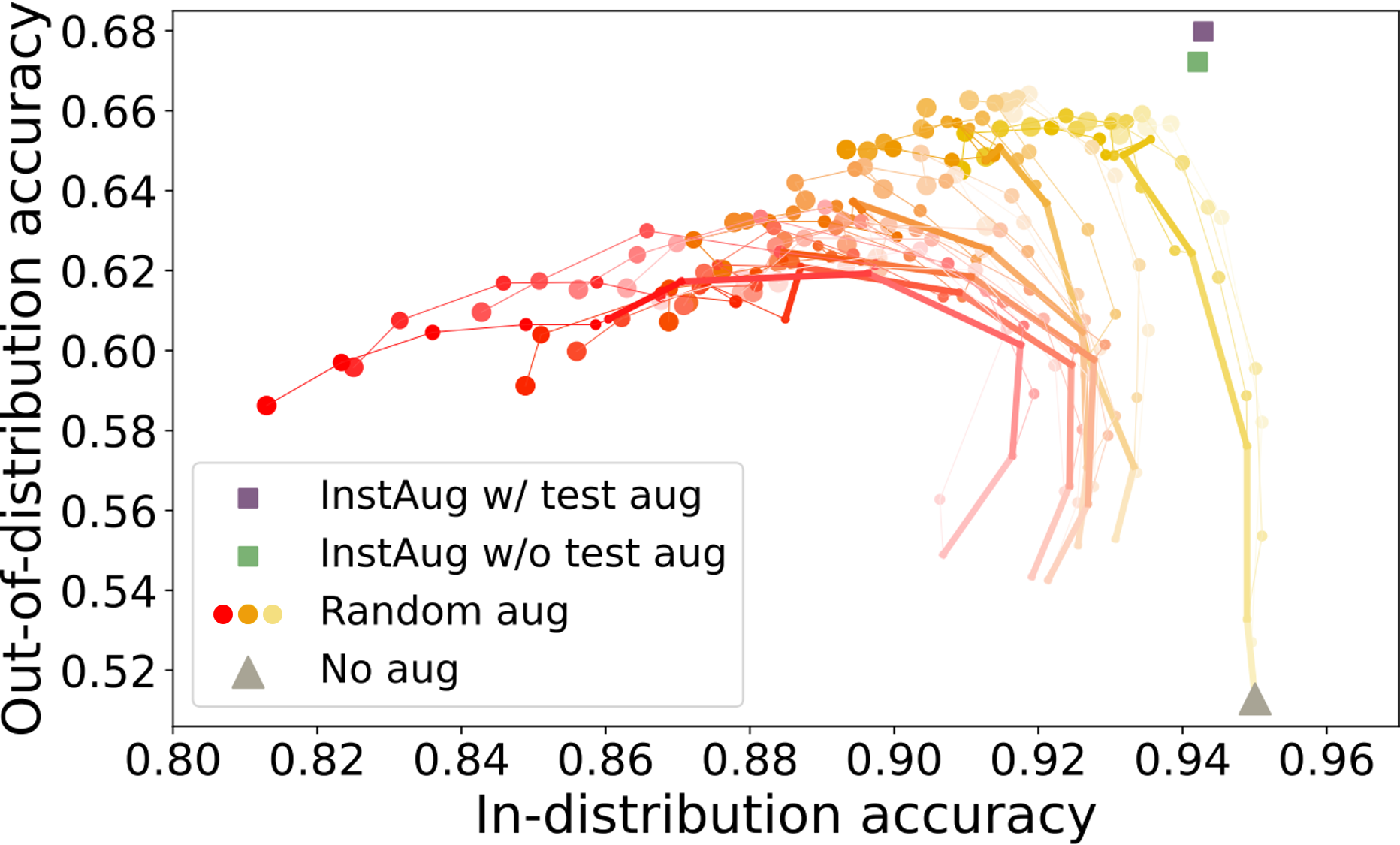}
 \caption{In- and out-of-distribution test accuracy for models trained on RawFooT D45. The round dots are random augmentation with different hyperparameter settings. The colors of dots change from yellow to red as hue jittering increases; more saturated dots indicate higher saturation jittering; larger dots mean higher brightness jittering. Thick lines connect dots with the same hue and brightness jitter, thin lines link dots with the same hue and saturation jitter.
 }
 \label{fig:color_2group_result}
  \vspace{-7pt}
\end{figure}
We also compare in-distribution and out-of-distribution generalization by splitting the 46 test sets into two groups, according to the similarity of their lighting conditions to D45---see \cref{apd:exp_details_rawfoot} for the details on the splitting method.
In \Cref{fig:color_2group_result} we can see that above a certain in-distribution performance, there exists a trade-off for random augmentation between in-distribution accuracy and out-of-distribution generalization, controlled through the hyperparameter settings. \method, meanwhile, delivers higher out-of-distribution performance than any of the hyperparameter configurations, while also simultaneously giving better in-distribution accuracy to the vast majority of them as well.

We can further vary the difficulty of the classification task by using different numbers of lighting conditions in the training data. In \cref{tab:color_difficulty_result}, we randomly select a set number of lighting conditions to use as the training set for each baseline.
As expected, the accuracy increases with the number of lighting conditions for all methods. However, the effect of random augmentation saturates: it performs similarly to no augmentation with 8 lighting conditions. By contrast, \method always provides improvements.
In \cref{apd:exp_details}, we show that these gains come at very little computational overhead at both train and test time.

\section{\method for Contrastive Learning}
Contrastive learning aims to learn features that are approximately invariant to certain augmentations. Typical contrastive learning methods, such as SimCLR~\citep{chen2020simple, ermolov2021whitening}, first sample two independent transformations, $\tau_1,\tau_2 \sim p(\tau)$, and apply them to an input image $\x$, generating two views $\x_1$ and $\x_2$. They then feed the transformed images to a neural encoder $f$, which is trained to maximize the similarity between $f(\x_1)$ and $f(\x_2)$, measured with a contrastive loss.

The choice of augmentations directly influences the learned invariance of the encoder and thus forms a crucial ingredient of contrastive learning~\citep{bachman2019learning,chen2020simple,tian2020makes}. 
Existing schemes use global augmentations that often introduce unrealistic assumptions. 
For example, if there are multiple entities in an image, such as grass and cattle in \cref{fig:idea_crop}, random cropping will pull features for different entities closer to each other. 
Consequently, we propose \method as a more flexible instance-specific augmentation method for contrastive learning.

Applying \method to contrastive learning is similar to the supervised case shown in \cref{sec:method}. The main difference is, given an input $\x$, we sample two $\tau$ independently from the input-specific distribution $p(\tau;\phi(\x))$, before they are applied to $\x$.
The training objective is correspondingly changed to minimizing the contrastive loss while keeping the diversity in a reasonable range.

We again consider TinyIN and evaluate three methods: \method, \method (without input), and Random crop. We exclude methods with uniform parameterization because of their earlier poor performance and note that AdaAug is not applicable for unsupervised learning.
All experiments are based on the SimCLR framework and use the PreActResNet-18 network as the encoder. We train each model with a batch size of 512 for 500 epochs. 
We then train a linear classifier to evaluate feature quality.
We use test-time augmentation---with 10 sampled crops---as this has been shown to improve performance~\citep{foster2021improving}.

\begin{figure}
 \centering
 \subfloat[]{
 \centering
 \includegraphics[width=0.13\textwidth]{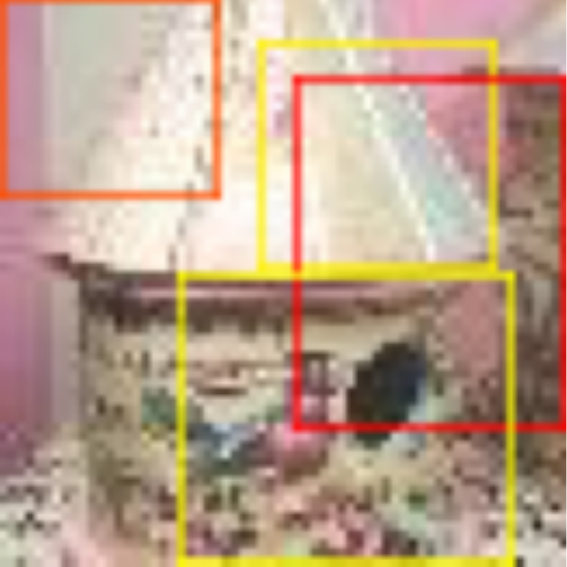}
 \label{fig:contrastive_0}
 }
 \subfloat[]{
 \centering
 \includegraphics[width=0.13\textwidth]{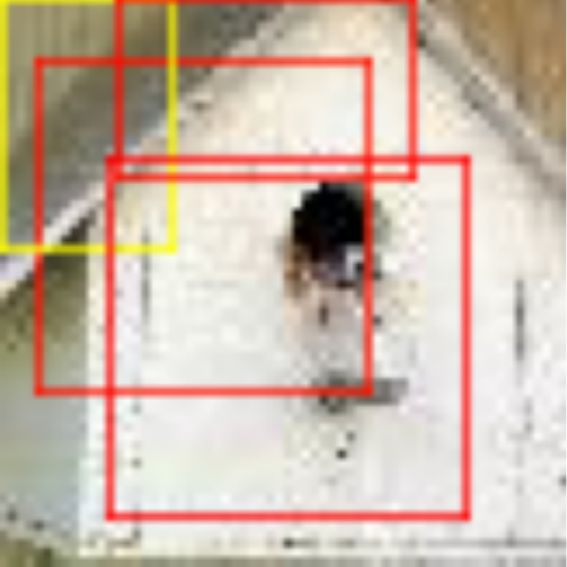}
 \label{fig:contrastive_1}
 }
 \subfloat[]{
 \centering
 \includegraphics[width=0.13\textwidth]{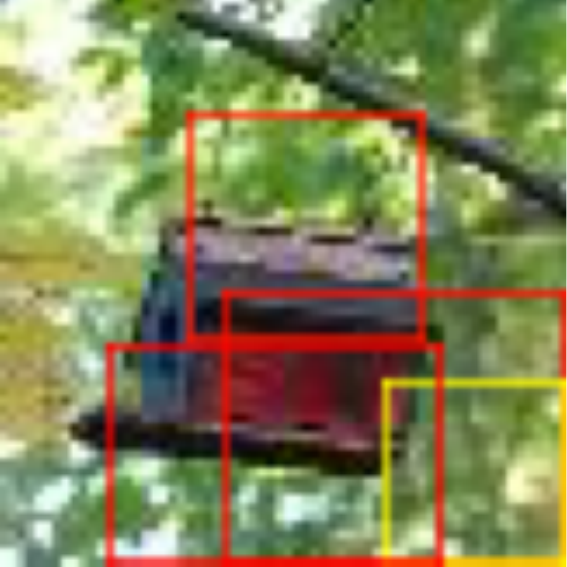}
 \label{fig:contrastive_2}
 }
 \caption{Examples of learned croppings by InstaAug for contrastive learning.  Conventions as per~\cref{fig:crop_examples}.
 \vspace{-12pt}
 }
 \label{fig:contrastive}
\end{figure}

\vspace{4pt}
\begin{table}
\scriptsize
    \centering
\caption{Representations learned by \method perform better in the downstream linear classification task than baselines. ${}^*$Results of Un-Mix are directly taken from \cite{shen2022mix}, which has the same network structure~(ResNet-18),  training algorithms~(SimCLR) and linear classifier as ours.
}
\begin{tabular}{ lc } 
 \toprule   
 Method\ & Accuracy (\%)\
 \\ \midrule
  
  Un-Mix~{\citep{shen2022mix}}         & ${49.58}^{*}$ \\
  Random crop         & ${51.63}_{\pm 0.30}$  \\
  
  \method~(without input)         & ${54.20}_{\pm 0.23}$ \\
  \method               & ${\bm{55.05}}_{\pm 0.21}$ \\
 \bottomrule
\end{tabular}
    \label{tab:contrastive_result}
\end{table}
\cref{tab:contrastive_result} shows that \method outperforms the random and global augmentation schemes as well as Un-Mix~\citep{shen2022mix}, which is a recent variant of MixUp methods for contrastive learning.
We see from the examples in \cref{fig:contrastive} that \method focuses on the salient features containing important information. 
We also notice that the sizes of learned patches are correlated to the sizes of the objects in the images.
Thus, \method is able to learn sensible instance-specific augmentations in a fully unsupervised setting.

\section{Conclusions}
\label{sec:discussion}
In this paper, we introduced \method, a method for learning instance-specific data augmentations that capture local invariances of the underlying data-generating process.
This is achieved by training an augmentation module that parametrizes an input-dependent distribution over transformations, whose samples can be used to augment the training data on the fly and/or for test-time augmentation.
The main benefits of \method stem from its applicability to a wide range of settings, its ease of use, and crucially its capacity to learn meaningful augmentations that in turn improve performance.
Empirically, we have demonstrated these benefits for both classification and contrastive learning problems, considering
several classes of transformations.

\section*{Acknowledgements}
We would love to thank Michael Hutchinson, Bryn Elesedy, and Sheheryar Zaidi for valuable discussions. We would also like to thank Google Cloud Platform (GCP) for their generous help in computing resources. Ning Miao gratefully acknowledges funding from China Scholarship Council - University of Oxford Scholarship as well as Tencent AI Labs through the Oxford-Tencent Collaboration on Large Scale Machine Learning.


\bibliography{main}
\bibliographystyle{icml2023}


\clearpage
\appendix
\numberwithin{equation}{section}
\numberwithin{figure}{section}
\numberwithin{table}{section}

\begin{appendices}

\section{Theoretical Analysis of Generalization Error}
\label{sec:app:theory}


We now provide a decomposition of the generalization error---i.e.~the difference between the true risk and the training risk---when using $\phi$ during training of the downstream classifier $f$.
Here we can view the objective of augmentation as adjusting the training objective to encourage the learned model to have a low true risk.
As such, the generalization error provides a measure of the effectiveness of the augmentation for the training of $f$;
by analyzing the behavior of the generalization error as a function of the augmentation module, we can derive a characterization of the desirable properties of the latter.

\newcommand{\ptrue}{p_{\mathrm{true}}}
\newcommand{\loss}{\mathcal{L}}
\newcommand{\Remp}{\hat{R}}
\newcommand{\Rt}{R_{\mathrm{tta}}}
\newcommand{\Yorg}{\hat{Y}}
\newcommand{\Ynew}{\tilde{Y}}
\newcommand{\disteq}{\overset{\mathrm{d}}{=}}
\newcommand{\termA}{\mathrm{(A)}}
\newcommand{\termB}{\mathrm{(B)}}
\newcommand{\termC}{\mathrm{(C)}}
\newcommand{\termBtta}{\mathrm{(B_{tta})}}

To start our analysis, we first define the true risk of the downstream model, $f$, as
\begin{align}
    R(f) := \E [\loss (f(X),Y)]
\end{align}
where $(X,Y)\sim \ptrue(X,Y)$ are drawn from the true data generating distribution.
In practice, one might also perform test-time augmentation, implying a different predictive function and thus different true risk, but for the purposes of our analysis, we will assume that this is not done, as this allows us to focus on the impact the invariance module has on $f$ during training.

On the other hand, the implied training risk (i.e.~our objective for training $f$) when using an invariance module is the augmented empirical risk
\begin{align}
\label{eq:Remp}
    \Remp(f,\phi) := \E [\loss(f(\tau(x_i)),y_i)]
\end{align}
where $i \sim \mathrm{Uniform}\{1,\dots,N\}$ is a uniformly sampled index for a point in the original training dataset $\{x_n,y_n\}_{n=1}^N$ and $\tau | i \sim p(\tau ; \phi(x_i))$ is the sampled transformation.
Note that the expectation in~\Cref{eq:Remp} is only over $i$ and $\tau$, with the data-points themselves not considered random variables for our purposes, because we are only provided with a single fixed training dataset.
%

The generalization error can now be defined as $\Remp(f,\phi)-R(f)$.
At a high level, we are interested in finding a $\phi$ that ensures this has a low magnitude.
More precisely, we want $\phi$ to ensure that the minimizer of the training risk, $\hat{f}^* :=\argmin_f \Remp(f,\phi)$, gives as low a true risk, $R(\hat{f}^*)$, as possible.
Therefore, we want to keep the generalization error magnitude small across different $f$ (relative to the corresponding variations in $\Remp(f,\phi)$ itself), so that the optima of the training and true risks are as similar as possible.
%
In other words, we want a $\phi$ that ensures $\Remp(f,\phi)-R(f)$ is small (in magnitude) for \emph{all} $f$, especially those close to $\hat{f}^*$.
If we do hypothetically drive the generalization error to zero for all $f$, we will have a mechanism for directly training to the true risk using a finite original training dataset.

To aid with decomposing the generalization error, it is convenient to further define the following random variables through their conditional distributions:
\begin{align}
    \Yorg | i &\sim \ptrue(Y=\Yorg|X=x_i) \quad \text{with} \quad \Yorg \indep \tau,  \\
    \Ynew | i, \tau &\sim \ptrue(Y=\Ynew | X=\tau(x_i)).
\end{align}
We can now write down our decomposition as follows:
\begin{align}
\begin{split}
\Remp(f,\phi) &- R(f)= {\color{olive}\underbrace{\E [\loss(f(\tau(x_i)),\Yorg)-\loss(f(\tau(x_i)),\Ynew)]}_{\termA}} \\
    &+ {\color{cyan} \underbrace{\E [\loss(f(\tau(x_i)),\Ynew)-\loss(f(X),Y)]}_{\termB}} \\
    &+ {\color{violet}\underbrace{\E [\loss(f(\tau(x_i)),y_i)-\loss(f(\tau(x_i)),\Yorg)]}_{\termC}}.
\end{split}
\end{align}
From this, we see that if the magnitude of {\color{olive}$\termA$}, {\color{cyan} $\termB$}, and {\color{violet}$\termC$} are all small, then our generalization error magnitude will be small as well.
Moreover, if we can construct a $\phi$ such that these terms are small for \emph{all} $f$, then we can ensure effective generalization performance.
%
We will now look at each term individually.


{\color{olive}$\termA$} provides a precise characterization of how well our transformation preserves the label distribution; it is the difference between the expected loss under the true label distribution of the untransformed inputs and the expected loss under the true label distribution of the transformed inputs,  making predictions using the transformed inputs in both cases.
In particular, by noting that we have
\begin{align}
{\color{olive}\termA} = \E \left[\E \left[\loss(f(\tau(x_i)),\Yorg)-\loss(f(\tau(x_i)),\Ynew)\middle| i, \tau\right]\right]
\end{align}
where $f(\tau(x_i))$ is deterministic given $\tau$ and $i$, we have that $\Ynew | i, \tau,  \disteq \Yorg | i, \forall i, \tau$ is a sufficient (but not necessary) condition to ensure ${\color{olive}}\termA=0$ for all $f$.\footnote{Note that $\Yorg \disteq \Ynew$ alone is not generally sufficient, as matching in marginal distribution does not ensure that the joint distributions with $i$ and $\tau$ also match, in turn yielding different expectations.}
That is, it is zero for all $f$ if the conditional distribution on the labels is the same for both the original and transformed inputs for all possible pairs $(i, \tau)$, i.e.~all possible original inputs and sampled transformations.
One simple way to ensure this is to have $\tau$ always be equal to the identity mapping, so this term prefers limited transformations.

By contrast, if the transformation destroys information about the label, $\Yorg | i$ and $\Ynew | i, \tau$ will now differ, such that, in general, ${\color{olive}\termA} \neq 0$ and, moreover, it will vary with $f$.
Here we typically expect that ${\color{olive}\termA} \ge 0$,\footnote{Note, though, that this is not formally guaranteed, even for the cross entropy loss and an $f$ that exactly captures the true distribution.  This is because, while Gibbs' inequality ensures the optimal $q$ given $p$ for a cross-validation expected loss $\E_{p(Y)}[-\log q(Y)]$ is $q=p$, in general, the optimal $p$ given $q$ is not $p=q$.} as we are making predictions using the transformed inputs, so the expected loss under the true label distribution for the transformed inputs will tend to be less than that when labels are generated using the untransformed input.
To keep the magnitude of ${\color{olive}\termA}$ low, we need to ensure that transformations maintain the conditional label distribution as well as possible, i.e.~that transformations preserve all input information that is salient for predicting labels.

Conveniently, minimizing $\Remp(f,\phi)$ with respect to $\phi$, as done by the \method training setup of~\Cref{sec:method_learn}, will naturally try to reduce ${\color{olive}\termA}$.
Given we expect the term to typically be positive, this provides an explanation for why \method can be effective without any separate consideration in the objective for the need for transformations to maintain the class label distribution.
%

${\color{cyan}\termB}$ represents how well our transformation captures the true input distribution.
Here we can utilize the fact that, by the definition of $\Ynew$,
\begin{align}
\begin{split}
    \E \left[\loss(f(\tau(x_i)),\Ynew) \middle| \tau(x_i) = x\right] &= \E \left[\loss(f(X),Y) \middle| X = x\right] \\
    &=: r(x)
\end{split}
\end{align}
to write it as
\begin{align}
    {\color{cyan}\termB} = \E [r(\tau(x_i))] - \E [r(X)],
\end{align}
where $r : \mathcal{X} \mapsto \mathbb{R}^+$ maps inputs to their true expected loss.
We thus see that $\tau(x_i) \disteq X$ is a sufficient (but not necessary) condition to ensure that ${\color{cyan}\termB} = 0$ for all $f$.
That is ${\color{cyan}\termB}$ is always $0$ if the process of choosing one of the training inputs at random followed by applying a sampled transformation to that input produces samples distributed exactly according to the true input distribution.
Unlike for ${\color{olive}\termA}$, there is no simple scenario in which we can ensure this is true, with the use of the identity transformation now likely to give significant discrepancies by failing to provide sufficient coverage of the input space: though the $x_i$ may originally have been sampled from $\ptrue(X)$, there is only a finite set of them, such that repeated sampling from this finite set represents a substantially different distribution to $\ptrue(X)$.
In fact, ${\color{cyan}\termB}$ nicely encapsulates the desire to perform augmentation in the first place, by showing how it can be used to increase the coverage of the input space.

How to best manage Term ${\color{cyan}\termB}$ will vary depending on the type of model used and the form of our transformations.
In some situations, it may be that no matter how diverse our transformations are within the class of those allowable, $\tau(x_i)$ will still only cover a subset of the support of $X$.
Here the most important factor for keeping ${\color{cyan}\termB}$ small will be to maximize the diversity of the transformations, e.g.~by maximizing their entropy, to ensure the best possible coverage of the true input space.
In other cases, it might also be possible to ``over--diversify'' the inputs, such that $\tau(x_i)$ can become more diffuse than $X$ for some choices of $\phi$, potentially causing training to lack focus on the particular test-time input distribution we care about.
Here we may need to ensure that the entropy of the transformation does not become so large as to cause such over-diversification, creating a more complex trade-off with the need to ensure sufficient coverage.
%
These two scenarios respectively motivate the lower and upper bounds on the transformation distribution entropy used when training the augmentation module.\footnote{Note here that the entropy bounds in~\Cref{sec:method_learn} are on are on the entropy on the parameters of $\tau$, rather than $\tau(x_i)$ itself.  This is because it is difficult to directly control the latter during the training, with the former providing a more practical proxy that is expected to generally be representative.}

For augmentation of high-dimensional data, the former, coverage-limited, scenario is expected to be significantly more likely, as our original training data will generally provide quite poor coverage of the true input distribution, while our transformations will not generally be sufficiently powerful to produce unrepresentative inputs.
Moreover, when working with large deep learning models, prediction in one region of the input space is rarely harmed by the addition of data in another input region.
Thus, for the typical scenarios, we expect \method to be deployed in, increasing the entropy of the transformations will directly relate to reducing the magnitude of ${\color{cyan}\termB}$.
Note here that it will typically be the case that ${\color{cyan}\termB} < 0$ provided that the transformations maintain the label distribution, as the accuracy of the downstream model will typically be higher for the transformations of the original training data that for the test data.

Term ${\color{violet}\termC}$ is the error from the fact that we only have one sample of the label for each original training input, rather than the full label distribution.
As $\Yorg \indep \tau$, we have limited ability to reduce it through controlling $\phi$; it essentially represents the irreducible noise in $\Remp(f,\phi)$ from only having a finite number of true labels.
Note that it is not related to the model's ability to generalize to unseen inputs, as it is based on variability in other possible labels we might have seen for our training inputs themselves; if $Y|X$ is actually deterministic, it is exactly zero.
%
As such, it is of limited interest for our analysis, while it will thankfully generally be much smaller than the other terms for practical problems unless we have both a very small dataset and a very noisy true label distribution.

Putting everything together, we see that ${\color{olive}\termA}$ and ${\color{cyan}\termB}$ respectively encapsulate the competing needs of the invariance module to maintain the conditional label distribution (i.e.~preserve the label information) and maximize coverage of the input space.
We have also seen that the former is typically naturally taken care of by minimizing $\Remp(f,\phi)$ with respect to $\phi$, motivating the cross-entropy term in~\Cref{eqn:main}, but the latter requires separate consideration, which we deal with through the regularization term.


\section{Details of Augerino}
\label{sec:ap:augerino}
As a method to learn invariance, Augerino~\citep{benton2020learning} is quite different from the previous approaches, which usually require an extra validation set. 
The basic idea behind Augerino is to use a few parameters~($\theta$) to control the transformation distribution on input images and learn these parameters with the training loss of the classifier.
Specifically, it minimizes the loss
\begin{subequations}
\begin{align}
\label{eqn:Augerino_loss}
\mathcal{L}_\lambda(\x;y) \triangleq  \mathbb{E} [\mathcal{L}(f(\tau(\x));y)] + \lambda \cdot \mathbb{R}(\theta),
\end{align}
\end{subequations}
where $\mathcal{L}(\x;y)$ is the cross-entropy loss and $\mathbb{R}(\theta)$ is a regularization function on the volume of the support of the distribution weighted by the hyper-parameter $\lambda$.

\paragraph{Comparison with \method.} \method shares with Augerino the ideas of tuning augmentation parameters by the classifier loss and using test time augmentation to boost performance, but they are different in the following aspects. The most significant difference is that \method is instance-specific, while Augerino learns global augmentations. 
Besides, Augerino uses a single scalar $\theta$ to parameterize a symmetric uniform distribution ($\mathcal{U}[-\theta, \theta]$) over each type of transformations, which lacks the flexibility to model more complex augmentations, such as cropping. 

In addition, Augerino uses a fixed weight $\lambda$ to balance the training loss and augmentation diversity. However, we find that, in more complicated settings, this is quite impractical. Specifically, we need different $\lambda$ in different stages of training. 
If we use a large $\lambda$ from the start of training, the diversity will quickly diverge to maximum, because the classifier is very weak and the loss is consequently dominated by the diversity term. This will block the training of the classifier because transformed samples from different classes are quite mixed with each other.
Otherwise, if we choose a small $\lambda$, the diversity will converge to zero after a few epochs, yielding similar results as the vanilla model without augmentation. In neither of the case can we learn a useful augmentation. 
Consequently, \method directly constrains the diversity to keep it stable during training.
\section{Method details}
\label{sec:app:method}
\subsection{Regression and self-supervised learning}
In \cref{sec:method}, we use classification as an example to introduce \method. However, \method can be easily applied to other tasks including regression and self-supervised learning. 
For regression, the classifier (see \cref{fig:model}) is replaced by a regressor and 
the loss function $\mathcal{L}$ in \cref{eqn:main} is changed accordingly to absolute or square error.
For self-supervised contrastive learning, we replace the classifier and cross-entropy loss with the feature extractor and contrastive loss (such as SimCLR loss~\citep{chen2020simple}), respectively. In addition, the sampler samples $2$ rather than $1$ transformations to generate multiple views for an input $\x$.

\vspace{-2pt}
\subsection{Test-time augmentation}
\vspace{-2pt}
\label{sec:test-time}
Besides augmenting data during training, the learned invariance can also be applied to test-time augmentation. Given a test image $\x$, we sample $n$ different transformations $\tau_i$ from $p(\tau; \phi(\x))$ and apply them to $\x$ to generate $n$ different views $\tau_i(\x)$. After feeding these views to the classifier, $f$, we use the mean logit $\frac{1}{n}\sum_{i=1}^n f(\tau_i(\x))$ to predict $\x$'s label.
When only learning invariance for test-time augmentation, \method can be trained with a fixed pre-trained classifier at a lower computation cost.

\vspace{-2pt}
\subsection{Other parametrization methods}
\vspace{-2pt}
Besides the uniform and location-related parameterization, we also tried VAE-like methods to parameterize augmentations, such as cropping. The main idea is to have a Gaussian latent variable and a neural decoder to map the latent Gaussian distributions to a continuous distribution on transformation parameters~(in this case, the centers and sizes of crops). 
However, similar to the uniform parameterization, we find the VAE-like parameterization unstable and easily stuck at local minima.

\begin{algorithm}[t]
\footnotesize
 \KwInput{Image $\x$, layer number $n\_layer$, channel number $M_i$}
 \KwOutput{Probability of patches $\mathbf{P}_{\text{crop}}$}
 
 $\mathbf{F_0^{'}}=\x$\;
 
 \For{$i = 1$; $i\leq$ n\_layer; $i=i+1$}{
   $\mathbf{F_i}$ = Conv2d($\mathbf{F^{'}_{i-1}}$, kernel=2, stride=1, output\_channel=$M_i$)
   
   $\mathbf{F_i^{'}}$ = Pooling($\mathbf{F_i}$, kernel=2)
   \tcp*{Conv and pooling}
   
   $\mathbf{F_i^{''}}$ = Conv2d($\mathbf{F_i^{'}}$, kernel=1, stride=1, output\_channel=1)
   \tcp{Concentrate info to single channel}
   
   $\mathbf{logit_i}$ = Flatten($\mathbf{F_i^{''}}$)
   \tcp*{Use activations of units at different layers as logits for patches of different sizes}
  }
   
 $\mathbf{logits}$ = Concat([$\mathbf{logit_i}$])
 
 $\mathbf{P}_{\text{crop}}$ = Normalize(Exponential($\mathbf{logits}$))
 \caption{Location related parameterization\label{alg:parameterization}}
 
\end{algorithm}

\vspace{-2pt}
\subsection{Network structures}
\vspace{-2pt}
In all of our experiments, we use PreActResNet-18 as the base structure of $\phi$ for both uniform and location-related parameterizations. When dealing with multiple transformations, for example in \cref{sec:exp_color}, we use the same network to generate parameters for all transformations simultaneously.

\newlength{\rawfoot}
\setlength{\rawfoot}{0.076\textwidth}

\begin{figure*}[t]
 \centering
\vspace{-5pt}
 \centering
 \setcounter{subfigure}{0}
 \subfloat{
 \centering
 \includegraphics[width=\rawfoot]{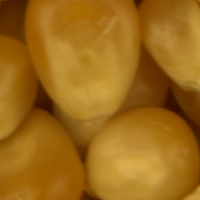}}
 \subfloat{
 \centering
 \includegraphics[width=\rawfoot]{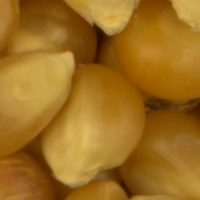}}
 \subfloat{
 \centering
 \includegraphics[width=\rawfoot]{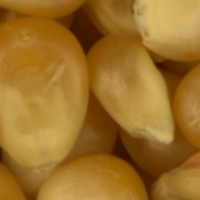}}
 \subfloat{
 \centering
 \includegraphics[width=\rawfoot]{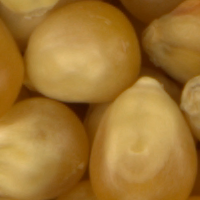}}
 \subfloat{
 \centering
 \includegraphics[width=\rawfoot]{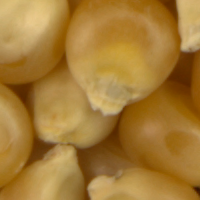}}
 \subfloat{
 \centering
 \includegraphics[width=\rawfoot]{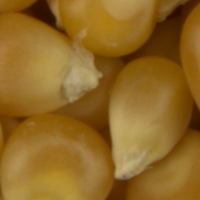}}
 \subfloat{
 \centering
 \includegraphics[width=\rawfoot]{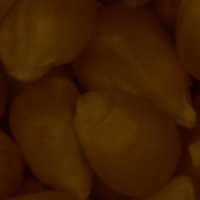}}
 \subfloat{
 \centering
 \includegraphics[width=\rawfoot]{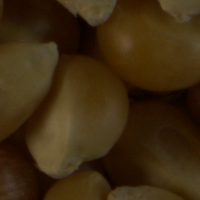}}
 \subfloat{
 \centering
 \includegraphics[width=\rawfoot]{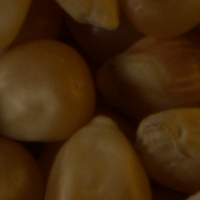}}
 \subfloat{
 \centering
 \includegraphics[width=\rawfoot]{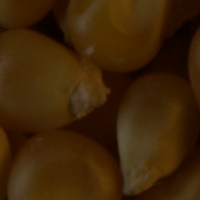}}
 \subfloat{
 \centering
 \includegraphics[width=\rawfoot]{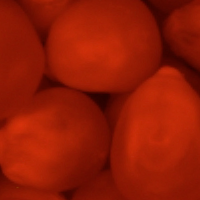}}
 \subfloat{
 \centering
 \includegraphics[width=\rawfoot]{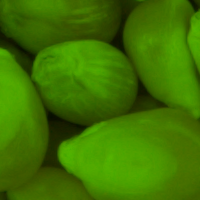}}

\vspace{-8pt}
 \centering
 \setcounter{subfigure}{0}
 \subfloat{
 \centering
 \includegraphics[width=\rawfoot]{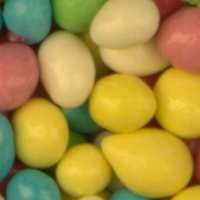}}
 \subfloat{
 \centering
 \includegraphics[width=\rawfoot]{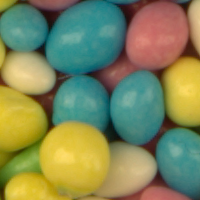}}
 \subfloat{
 \centering
 \includegraphics[width=\rawfoot]{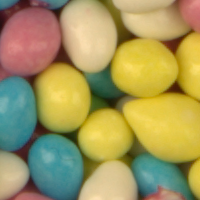}}
 \subfloat{
 \centering
 \includegraphics[width=\rawfoot]{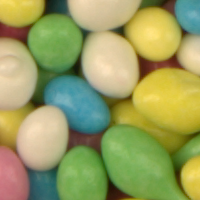}}
 \subfloat{
 \centering
 \includegraphics[width=\rawfoot]{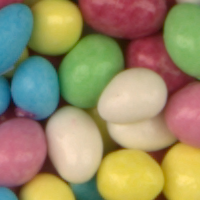}}
 \subfloat{
 \centering
 \includegraphics[width=\rawfoot]{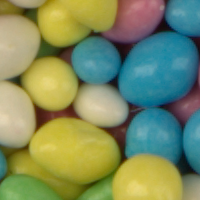}}
 \subfloat{
 \centering
 \includegraphics[width=\rawfoot]{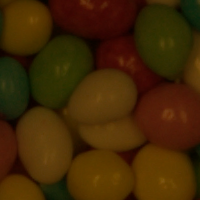}}
 \subfloat{
 \centering
 \includegraphics[width=\rawfoot]{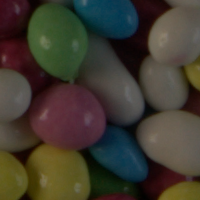}}
 \subfloat{
 \centering
 \includegraphics[width=\rawfoot]{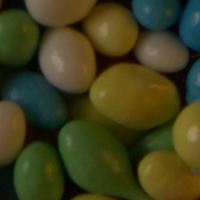}}
 \subfloat{
 \centering
 \includegraphics[width=\rawfoot]{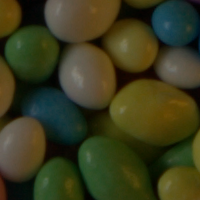}}
 \subfloat{
 \centering
 \includegraphics[width=\rawfoot]{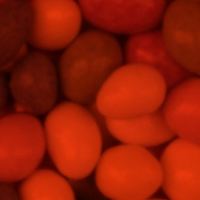}}
 \subfloat{
 \centering
 \includegraphics[width=\rawfoot]{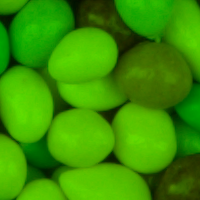}}

\vspace{-8pt}
 \centering
 \setcounter{subfigure}{0}
 \subfloat{
 \centering
 \includegraphics[width=\rawfoot]{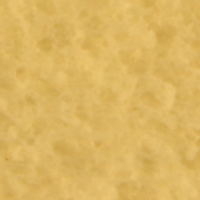}}
 \subfloat{
 \centering
 \includegraphics[width=\rawfoot]{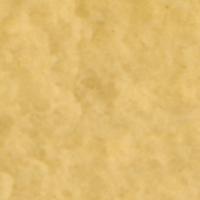}}
 \subfloat{
 \centering
 \includegraphics[width=\rawfoot]{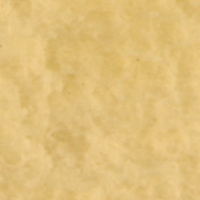}}
 \subfloat{
 \centering
 \includegraphics[width=\rawfoot]{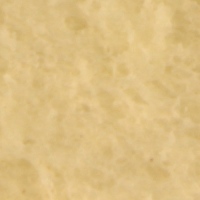}}
 \subfloat{
 \centering
 \includegraphics[width=\rawfoot]{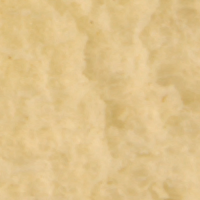}}
 \subfloat{
 \centering
 \includegraphics[width=\rawfoot]{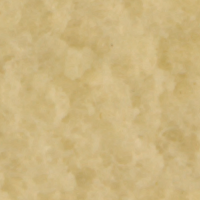}}
 \subfloat{
 \centering
 \includegraphics[width=\rawfoot]{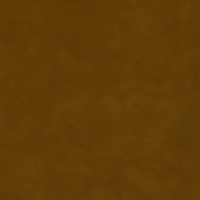}}
 \subfloat{
 \centering
 \includegraphics[width=\rawfoot]{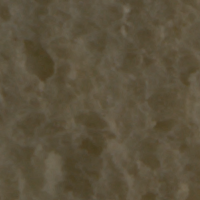}}
 \subfloat{
 \centering
 \includegraphics[width=\rawfoot]{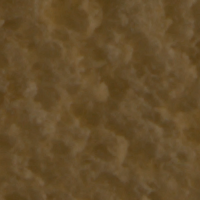}}
 \subfloat{
 \centering
 \includegraphics[width=\rawfoot]{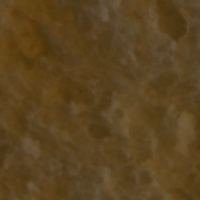}}
 \subfloat{
 \centering
 \includegraphics[width=\rawfoot]{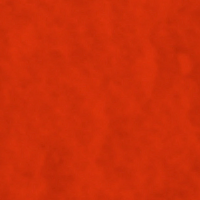}}
 \subfloat{
 \centering
 \includegraphics[width=\rawfoot]{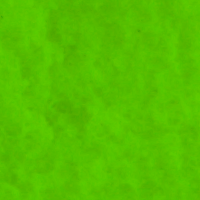}}

 \vspace{-8pt}
 \centering
 \setcounter{subfigure}{0}
 \subfloat{
 \centering
 \includegraphics[width=\rawfoot]{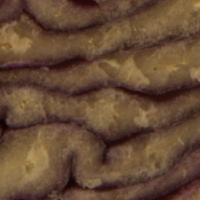}}
 \subfloat{
 \centering
 \includegraphics[width=\rawfoot]{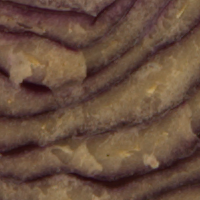}}
 \subfloat{
 \centering
 \includegraphics[width=\rawfoot]{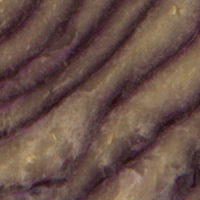}}
 \subfloat{
 \centering
 \includegraphics[width=\rawfoot]{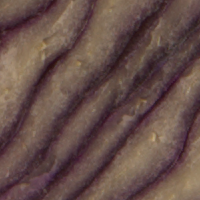}}
 \subfloat{
 \centering
 \includegraphics[width=\rawfoot]{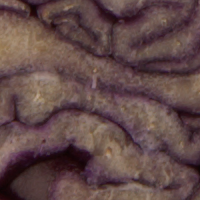}}
 \subfloat{
 \centering
 \includegraphics[width=\rawfoot]{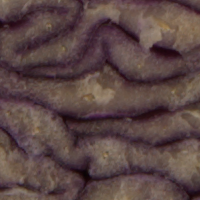}}
 \subfloat{
 \centering
 \includegraphics[width=\rawfoot]{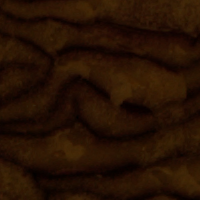}}
 \subfloat{
 \centering
 \includegraphics[width=\rawfoot]{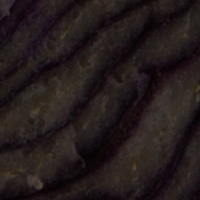}}
 \subfloat{
 \centering
 \includegraphics[width=\rawfoot]{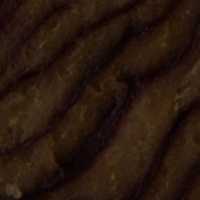}}
 \subfloat{
 \centering
 \includegraphics[width=\rawfoot]{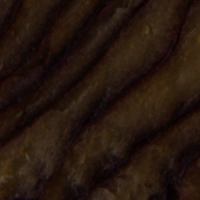}}
 \subfloat{
 \centering
 \includegraphics[width=\rawfoot]{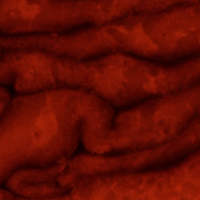}}
 \subfloat{
 \centering
 \includegraphics[width=\rawfoot]{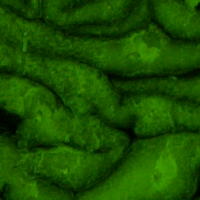}}
 
 \vspace{-8pt}
 \subfloat{
    \centering
    \begin{minipage}{\rawfoot}
      \centering
      \tiny D45
    \end{minipage}}
 \subfloat{
    \centering
    \begin{minipage}{\rawfoot}
      \centering
      \tiny D50
    \end{minipage}}
  \subfloat{
    \centering
    \begin{minipage}{\rawfoot}
      \centering
      \tiny D55
    \end{minipage}}
 \subfloat{
    \centering
    \begin{minipage}{\rawfoot}
      \centering
      \tiny D60
    \end{minipage}}
 \subfloat{
    \centering
    \begin{minipage}{\rawfoot}
      \centering
      \tiny D65~(I=1)
    \end{minipage}}
 \subfloat{
    \centering
    \begin{minipage}{\rawfoot}
      \centering
      \tiny D65~(I=0.75)
    \end{minipage}}
 \subfloat{
    \centering
    \begin{minipage}{\rawfoot}
      \centering
      \tiny D27~(A=$90$)
    \end{minipage}}
 \subfloat{
    \centering
    \begin{minipage}{\rawfoot}
      \centering
      \tiny D65+D95
    \end{minipage}}
 \subfloat{
    \centering
    \begin{minipage}{\rawfoot}
      \centering
      \tiny D65+D27
    \end{minipage}}
 \subfloat{
    \centering
    \begin{minipage}{\rawfoot}
      \centering
      \tiny D95+D27
    \end{minipage}}
 \subfloat{
    \centering
    \begin{minipage}{\rawfoot}
      \centering
      \tiny Red
    \end{minipage}}
 \subfloat{
    \centering
    \begin{minipage}{\rawfoot}
      \centering
      \tiny Green
    \end{minipage}}
 \vspace{-4pt}
 \caption{Examples of RawFooT data. Each row contains images in the same class~(corn, candies, floor, red cabbage) under different lighting conditions. The left and right half of lighting conditions are in the easy and hard groups, respectively.
 }
 \label{fig:rawfoot_dataset_example}
 \vspace{-10pt}
\end{figure*}

\section{Experimental details}
\label{apd:exp_details}

\subsection{Cropping}
\label{apd:exp_details_cropping}

\paragraph{Supervised training}
Based on the Mixmo codebase\footnote{\url{https://github.com/alexrame/mixmo-pytorch.git}, under Apache License v2.0.}~\citep{rame2021mixmo}, we use stochastic gradient descent~(SGD) optimizer to train baselines and \method. 
For the classifier, the initial learning rate is set to $0.2$ (with momentum $0.9$ and weight decay $1e-4$). 
A scheduler is used to decrease the learning rate by a factor of $0.9$ once validation accuracy doesn't increase for 10 epochs. The learning rate of the augmentation module $\phi$ is fixed at $1e-5$. Batch size is set to 100 and we pre-train \method for 10 epochs without augmentation. We train the model until convergence and the maximum epoch is set to $150$.

\paragraph{Contrastive training}
We directly apply \method on the codebase\footnote{\url{https://github.com/htdt/self-supervised.git}, under Apache License v2.0.} from \cite{ermolov2021whitening}. Because of the characteristics of contrastive learning, we set the batch size to $512$. 
Same as the supervised case, we use SGD optimizer to train the augmentation module $\phi$.
Differently, we use Adam optimizer~\citep{kingma2015adam} (with learning rate $1e-3$ and weight decay $1e-6$) to train the base model. We train each model for $500$ epochs and decrease the learning rate by a factor of $0.8$ at step $450$ and $475$.

\paragraph{Implementation of LRP}
As an example, we show how to implement location-related parameterization with a basic CNN structure in \cref{alg:parameterization}.

\subsection{Color jittering on textures} \label{apd:exp_details_rawfoot}

\paragraph{Training.}
We use PreActResNet-18 ($width=1$) on texture recognition task on RawFooT and train it with SGD optimizer. The learning rate is $0.02$ (with momentum $0.9$ and weight decay $1e-4$) for the classifier and $1e-5$ for the augmentation module $\phi$. We train each model for $50$ epochs and learning rate schedulers are not necessary in this task.

\paragraph{Random augmentation baseline.} We
sweep over the variation range on each channel to find the best hyperparameters for the random augmentation baseline. 
For hue~(h-jittering), we sweep between $[0, 0.5]$ with stride $0.1$, and for saturation~(s-jittering) as well as brightness value~(v-jittering), we sweep between $[0, 1.0]$ with stride $0.2$, which yields 216 different settings in total. The best accuracy shown in \cref{tab:color_overall_result} is achieved where h,s,v$=0.0, 0.2, 0.8$. 

\begin{table}[t]
    \centering
    \small
    \caption{Splitting of Lighting conditions.}
    \vspace{-3pt}
\begin{tabular}{ lc } 
 \toprule
 Group & Lighting id \\\midrule
 Easy~(1) &  1-4,10,14-31\\
 Hard~(2) & 5-9, 11-13, 32-46 \\
 \bottomrule
\end{tabular}
\label{tab:split}
\vspace{-12pt}
\end{table}

\paragraph{In-distribution vs. out-of-distribution generalization.}

To further investigate the effect of each augmentation method, we additionally split the 46 test sets into two equally-sized groups. 
The first group contains lighting conditions similar to D45, such as daylight with different temperatures, for which the vanilla model without augmentation trained on D45 has high test accuracy. 
The second group contains lighting conditions that are dramatically different from D45, for example, pure red light, which are more difficult for the vanilla method. 
Then the average accuracy on the first group can be regarded as a measure of in-distribution generalization, while the accuracy on the second group reflects out-of-distribution generalization. 

\newlength{\scalingwidth}
\setlength{\scalingwidth}{0.45\textwidth}

\begin{figure}
 \centering
 \vspace{-8pt}
 \subfloat{
    \centering
    \begin{minipage}{0.15\scalingwidth}
      \centering
      (\rom{1})~Input
    \end{minipage}}
  \hspace{0.00\scalingwidth}     
  \subfloat{
     \centering
     \begin{minipage}{0.57\scalingwidth}
       \center (\rom{2})~\method
      \end{minipage}}
    \hspace{0.00\scalingwidth}     
  \subfloat{
     \centering
     \begin{minipage}{0.23\scalingwidth}
       \center (\rom{3})~Augerino
      \end{minipage}}     
 \vspace{-8pt}
 
 \subfloat{
 \centering
 \includegraphics[width=0.16\scalingwidth]{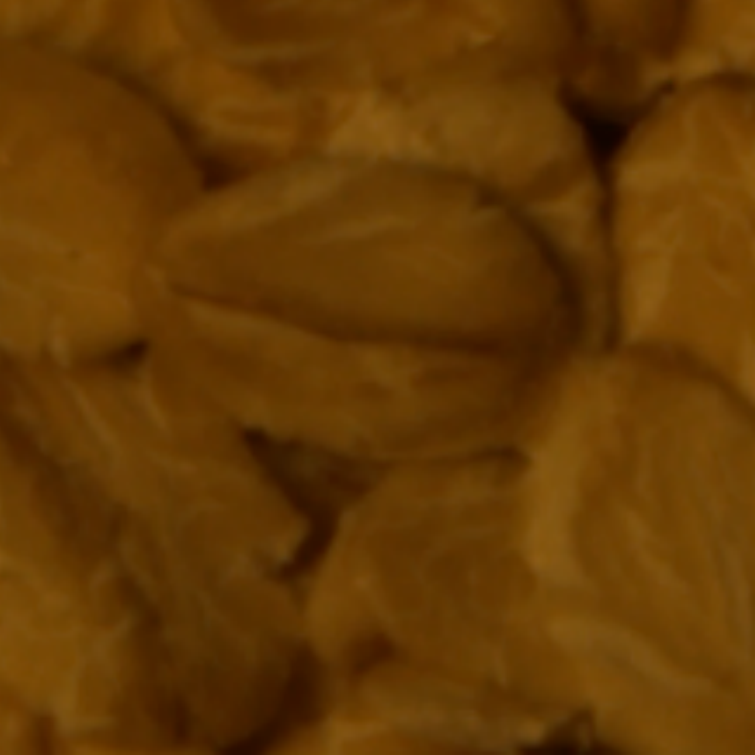}
 }
 \subfloat{
 \centering
 \includegraphics[width=0.10\scalingwidth]{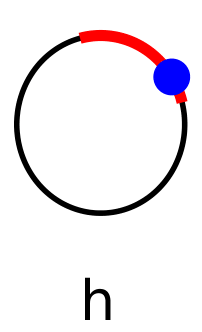}
 }
 \subfloat{
 \centering
 \includegraphics[width=0.09\scalingwidth]{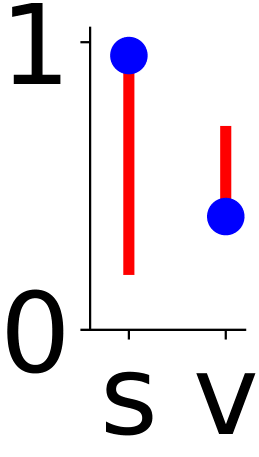}
 }
 \subfloat{
 \centering
 \includegraphics[width=0.16\scalingwidth]{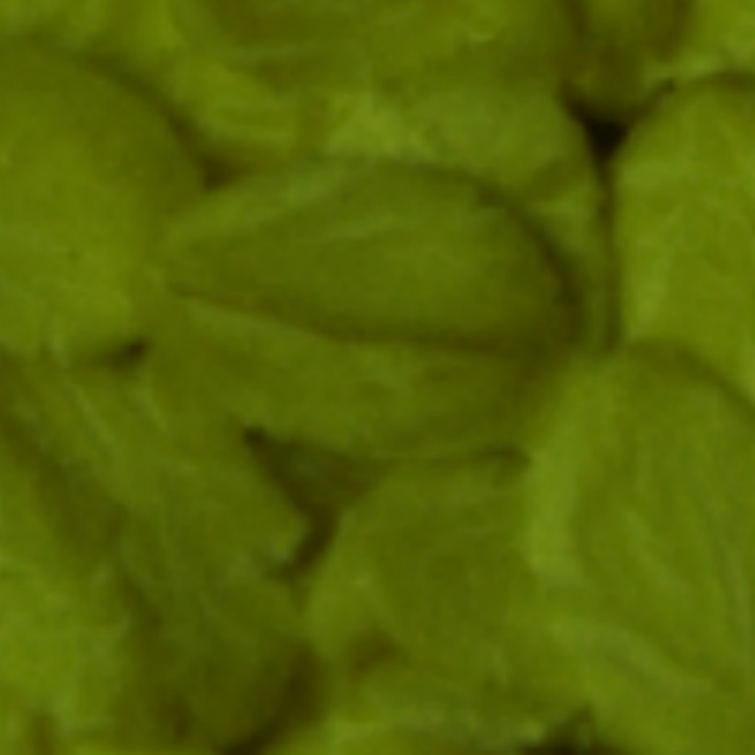}
 }
 \subfloat{
 \centering
 \includegraphics[width=0.16\scalingwidth]{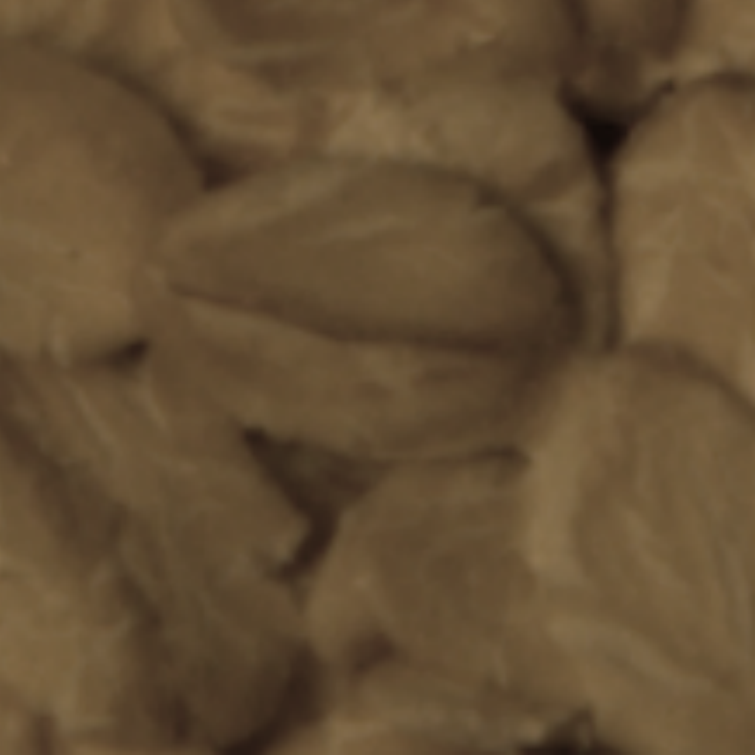}
 }
 \subfloat{
 \centering
 \includegraphics[width=0.10\scalingwidth]{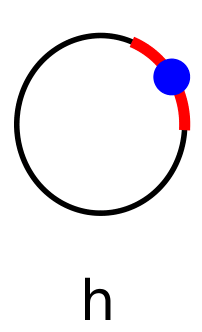}
 }
 \subfloat{
 \centering
 \includegraphics[width=0.09\scalingwidth]{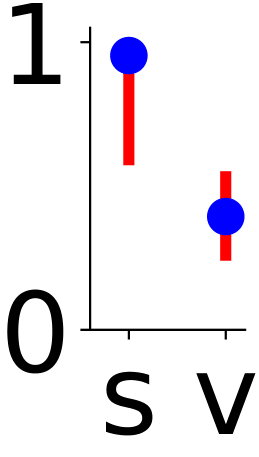}
 }
 
 \subfloat{
 \centering
 \includegraphics[width=0.16\scalingwidth]{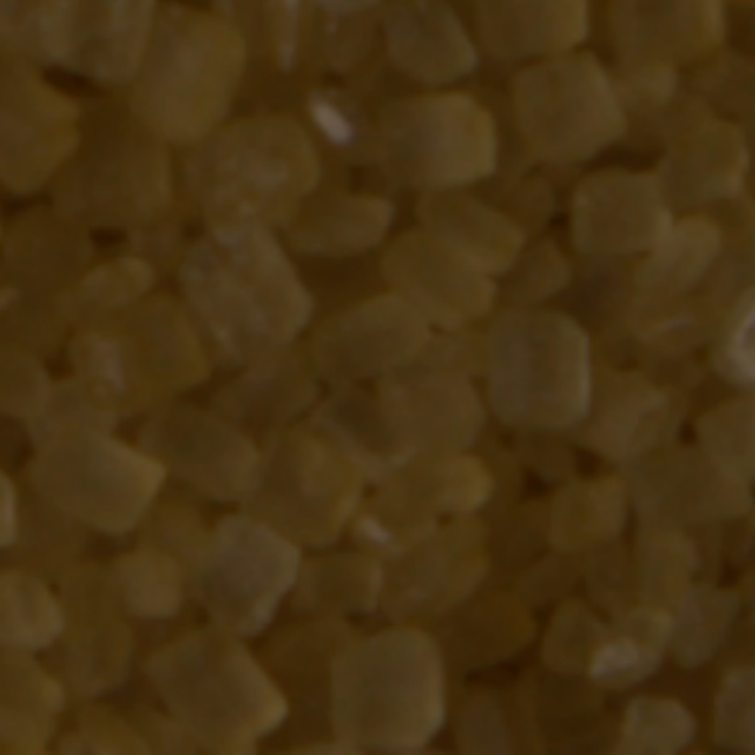}
 }
 \subfloat{
 \centering
 \includegraphics[width=0.10\scalingwidth]{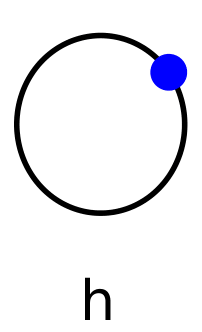}
 }
 \subfloat{
 \centering
 \includegraphics[width=0.09\scalingwidth]{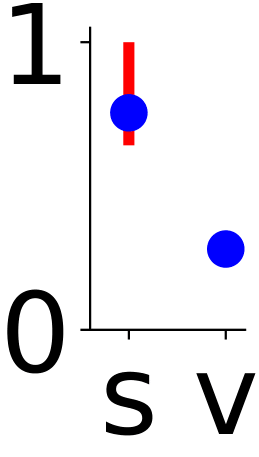}
 }
 \subfloat{
 \centering
 \includegraphics[width=0.16\scalingwidth]{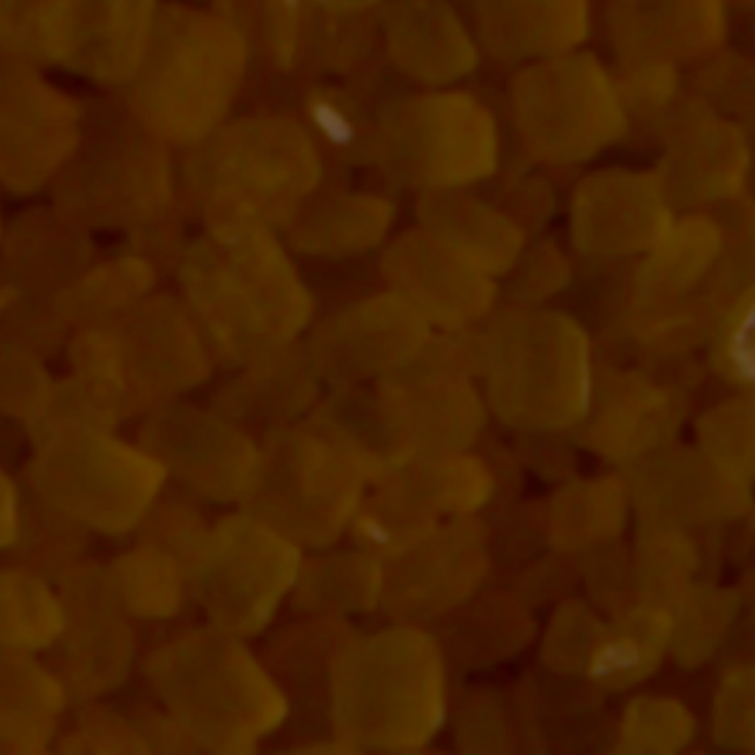}
 }
 \subfloat{
 \centering
 \includegraphics[width=0.16\scalingwidth]{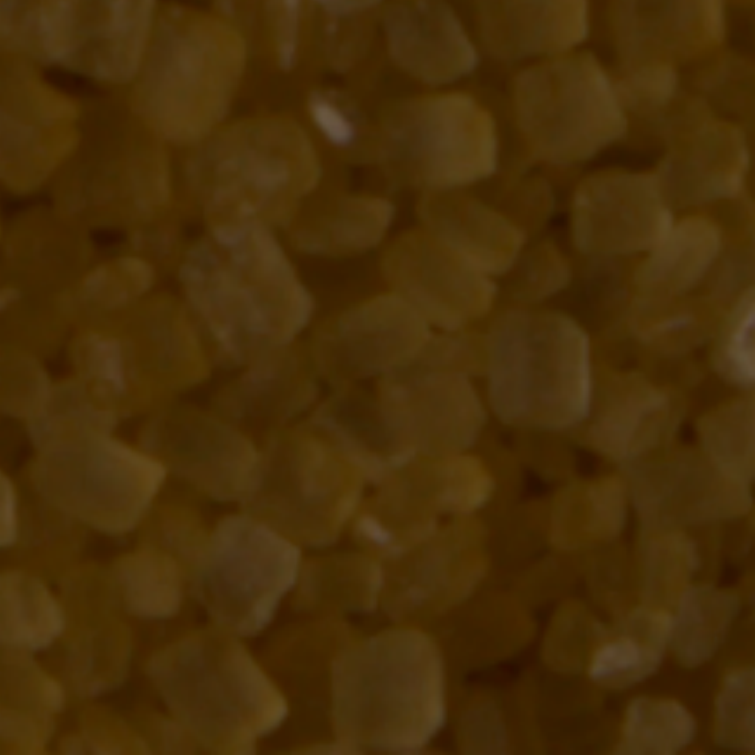}
 }
 \subfloat{
 \centering
 \includegraphics[width=0.10\scalingwidth]{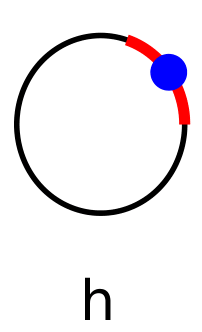}
 }
 \subfloat{
 \centering
 \includegraphics[width=0.09\scalingwidth]{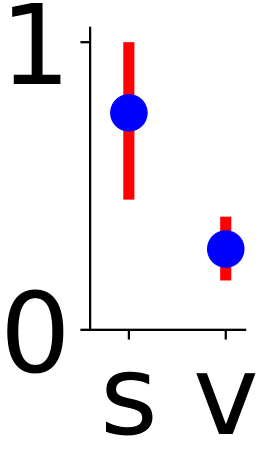}
 }
 
 \subfloat{
 \centering
 \includegraphics[width=0.16\scalingwidth]{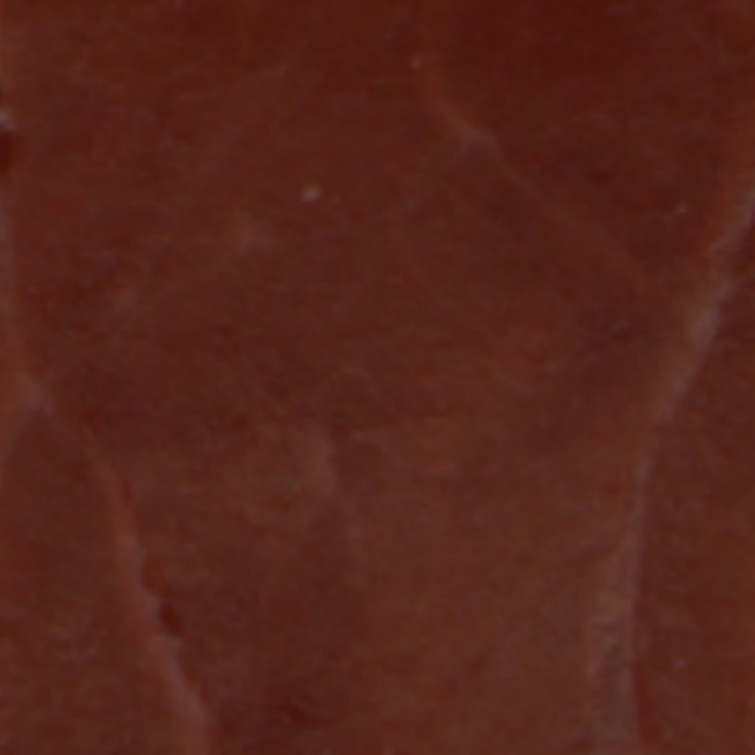}
 }
 \subfloat{
 \centering
 \includegraphics[width=0.10\scalingwidth]{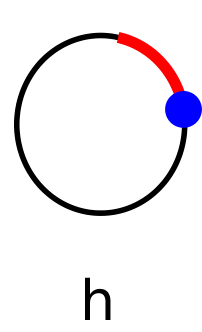}
 }
 \subfloat{
 \centering
 \includegraphics[width=0.09\scalingwidth]{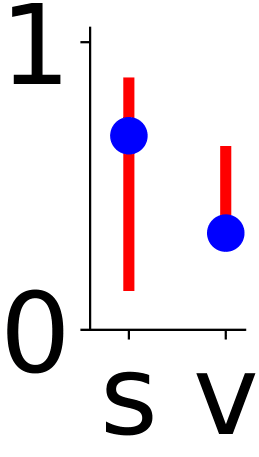}
 }
 \subfloat{
 \centering
 \includegraphics[width=0.16\scalingwidth]{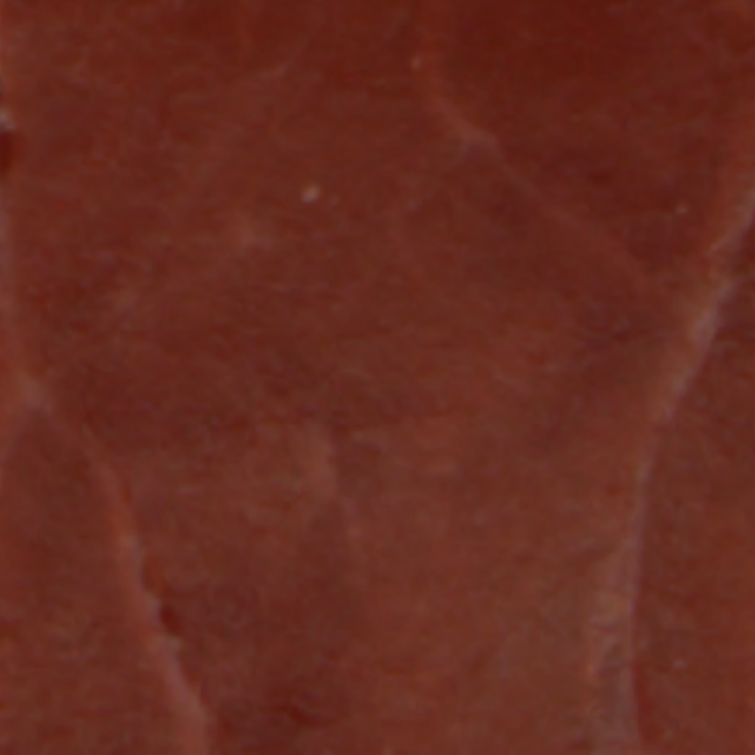}
 }
 \subfloat{
 \centering
 \includegraphics[width=0.16\scalingwidth]{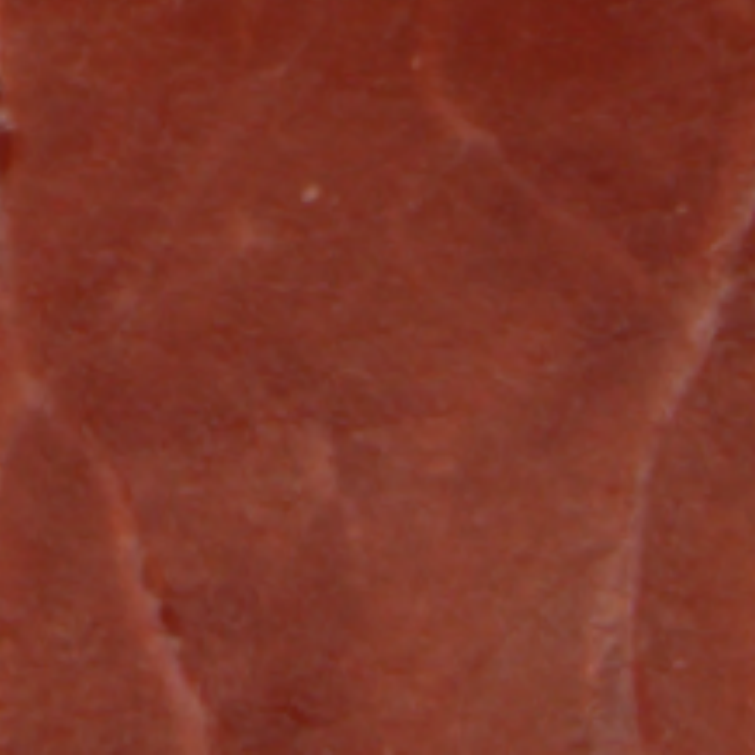}
 }
 \subfloat{
 \centering
 \includegraphics[width=0.11\scalingwidth]{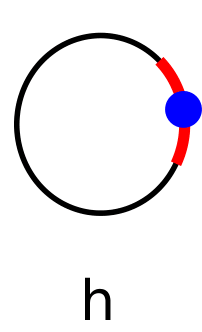}
 }
 \subfloat{
 \centering
 \includegraphics[width=0.09\scalingwidth]{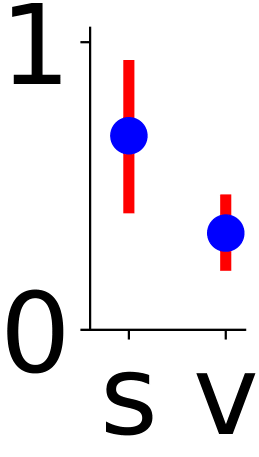}
 }
 
 \setcounter{subfigure}{0}
 \subfloat[]{
 \centering
 \includegraphics[width=0.16\scalingwidth]{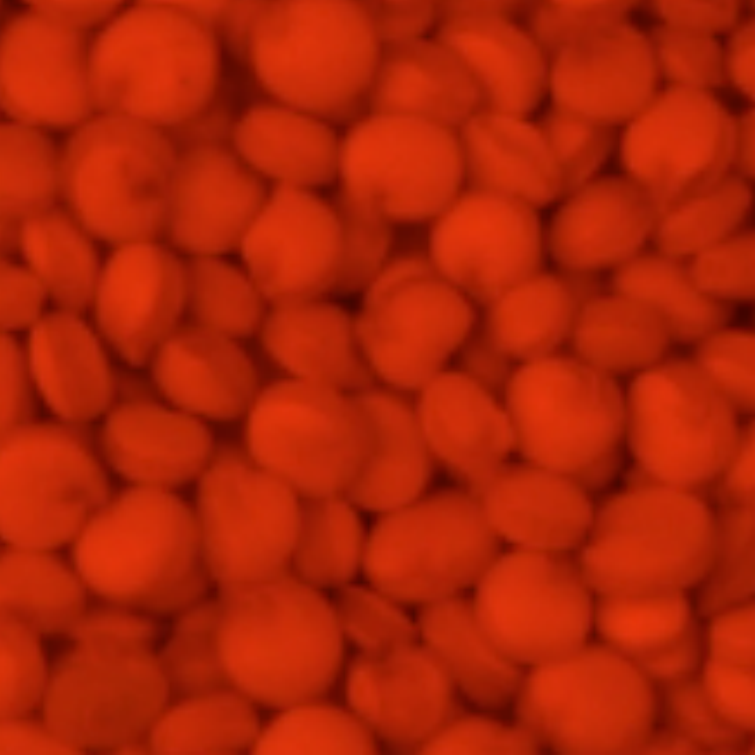}
 }
 \subfloat[]{
 \centering
 \includegraphics[width=0.10\scalingwidth]{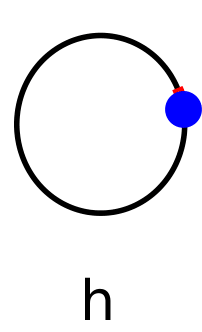}
 }
 \subfloat[]{
 \centering
 \includegraphics[width=0.09\scalingwidth]{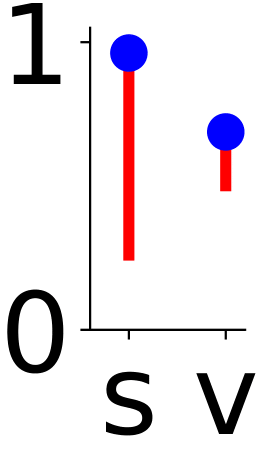}
 }
 \subfloat[]{
 \centering
 \includegraphics[width=0.16\scalingwidth]{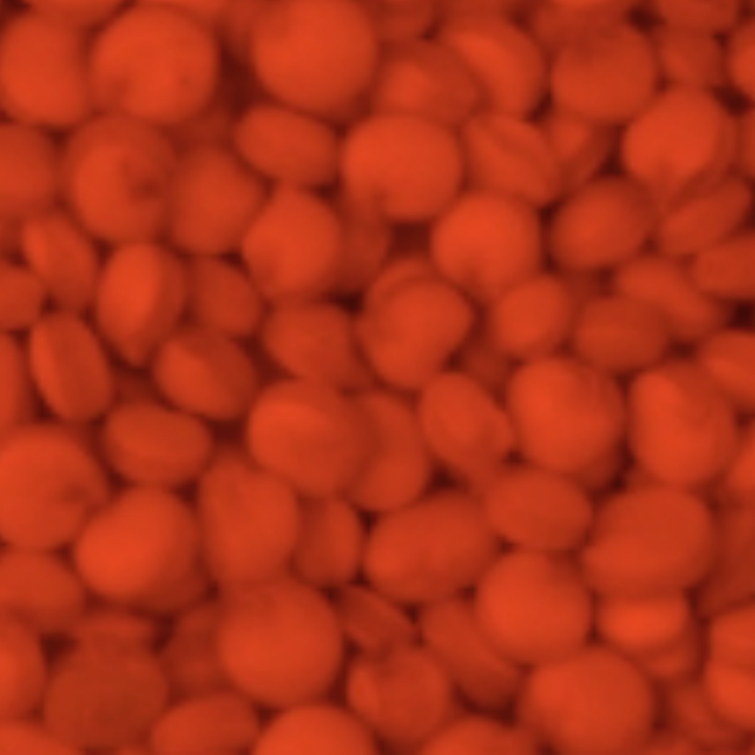}
 }
 \subfloat[]{
 \centering
 \includegraphics[width=0.16\scalingwidth]{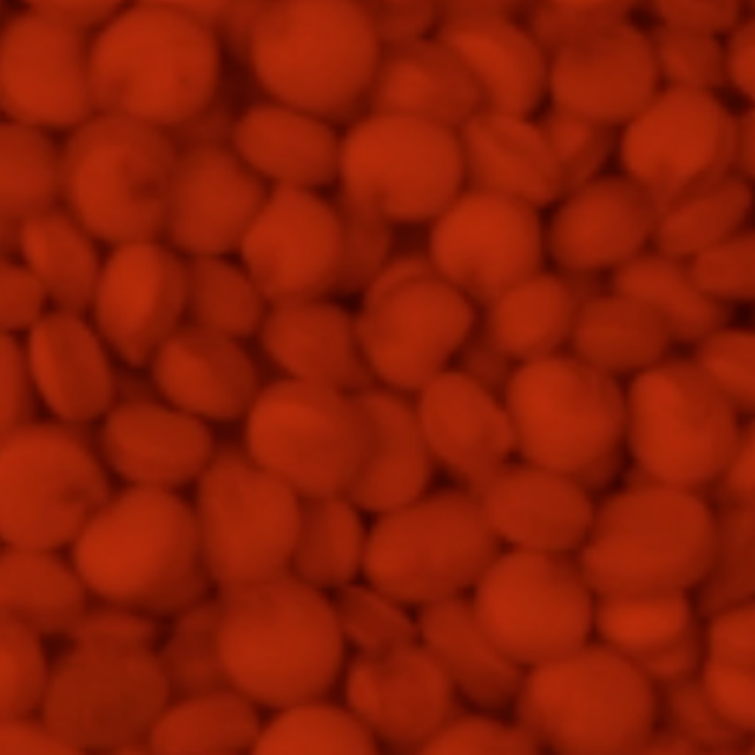}
 }
 \subfloat[]{
 \centering
 \includegraphics[width=0.10\scalingwidth]{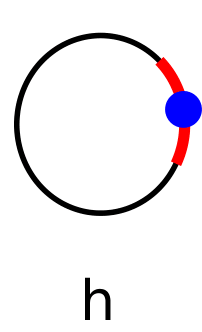}
 }
 \subfloat[]{
 \centering
 \includegraphics[width=0.09\scalingwidth]{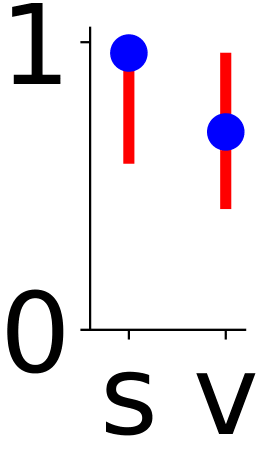}
 }
 \vspace{-6pt}
 \caption{Examples of learned color jittering. (a) Original image; (b, f) Average hue~(H) of original image~(blue dot) and learned hue jittering~(red arc) for \method and Augerino; (c,g) learned saturation (S) and brightness value (V) of original image~(blue dot) and learned hue jittering~(red line segment) for \method and Augerino; (d,e) examples of images transformed by \method. 
 }
 \label{fig:color_example}
\end{figure}

\subsection{Time complexity} On a single 1080Ti, each iteration of training InstaAug on TinyIN takes 0.25s, and each epoch takes 250s. As it takes 150 epochs (\num{1.5e5} iterations) for the model to converge, the total training time is about 10h. For random augmentation, each iteration takes 0.15s, and each epoch takes 150s. It takes the same 150 epochs (\num{1.5e5} iterations) to converge, so the total training time is about 7h. All of the other augmentation learning approaches are slower than random augmentation, with some of them being noticeably slower than InstaAug itself. For example, the exploitation stage of AdaAug has the time complexity as random augmentation, which is 0.15s/iter, 150s/epoch, and 7h for the whole training process. However, its exploration took us more than 15h because it requires averaging the representations of a large number of augmented samples.

The training speed of \method on RawFoot (color jittering) is similar to random augmentation (0.37s/iter vs. 0.40s/iter), though it takes more epochs (about 40) compared with random augmentation, which usually converges after 25 epochs.

\section{Additional Results and Discussion}
\label{apd:add_results}

\subsection{RawFooT}
\label{apd:add_results_rawfoot}

\cref{fig:color_example} shows some examples of learned color jittering. Though it's not easy to fully understand them, we can still find some patterns. For example, \method tends to increase the brightness of darker images~(row 1 and 3) and decrease the brightness of brighter images~(row 4). Also, \method is more likely to change saturation compared with hue and brightness, which is consistent with the common belief that saturation contains less information than hue and brightness.

\begin{table}

    \centering
    \vspace{7pt}
    \caption{
Model performance with different choice of $\text{H}_{\min}$ and $\text{H}_{\max}$ on supervised cropping.}
\begin{tabular}{ ccc } 
 \toprule   
 $\text{H}_{\min}$ &$\text{H}_{\max}$ & Accuracy (\%)\
 \\ \midrule
  $0.0$ &$0.5$&$52.12$ \\
  $0.5$ &$1.0$& $61.28$ \\
  $1.0$ &$1.5$& $62.91$ \\
  $1.5$ &$2.0$& $64.39$ \\
  $2.0$ &$2.5$& $65.04$ \\
  $2.5$ &$3.0$& $65.05$ \\
  $3.0$ &$3.5$& $\textbf{66.03}$ \\
  $3.5$ &$4.5$& $65.60$ \\
  $4.0$ &$4.5$& $64.35$ \\
  $4.5$ &$5.0$& $64.17$ \\\midrule
  $0.0$ &$1.0$& $51.78$ \\
  $1.0$ &$2.0$& $63.96$ \\
  $2.0$ &$3.0$& $65.25$ \\
  $3.0$ &$4.0$& $\textbf{65.78}$ \\
  $4.0$ &$5.0$& $64.23$ \\
 \bottomrule
\end{tabular}

    \label{apptab:H_ablation}
\end{table}
\method 's behavior is quite different on different samples. It even decides not to augment the H and V channels of the image in the second row.
In comparison, Augerino adds or multiplies noise to each channel with the same distribution across all samples, which is harmful in many cases. For example, the input image in the last row is already very bright. but Augerino allows further increasing its brightness. Then brightness values of many pixels will be capped at 1.0, which leads to loss of information.


\subsection{Hyperparameter Ablation}
\label{app:sec:ablation}

The two hyperparameters of \method are $\text{H}_{\min}$ and $\text{H}_{\max}$, which reflect human preference on augmentation diversity. To investigate how $\text{H}_{\min}$ and $\text{H}_{\max}$ influence model performance and provide a guide on how to choose them, we perform an ablation study for the experiment of~\Cref{sec:exp_cropping}, wherein we sweep over possible intervals of length $0.5$ and $1.0$. From \cref{apptab:H_ablation}, we find that the best accuracy is achieved when $[\text{H}_{\min}, \text{H}_{\max}]$ is set to $[3, 3.5]$, while any sub-interval of $[2, 4]$ produces significantly better results compared with random augmentation.

\begin{figure}
 \centering
 \includegraphics[width=0.45\textwidth]{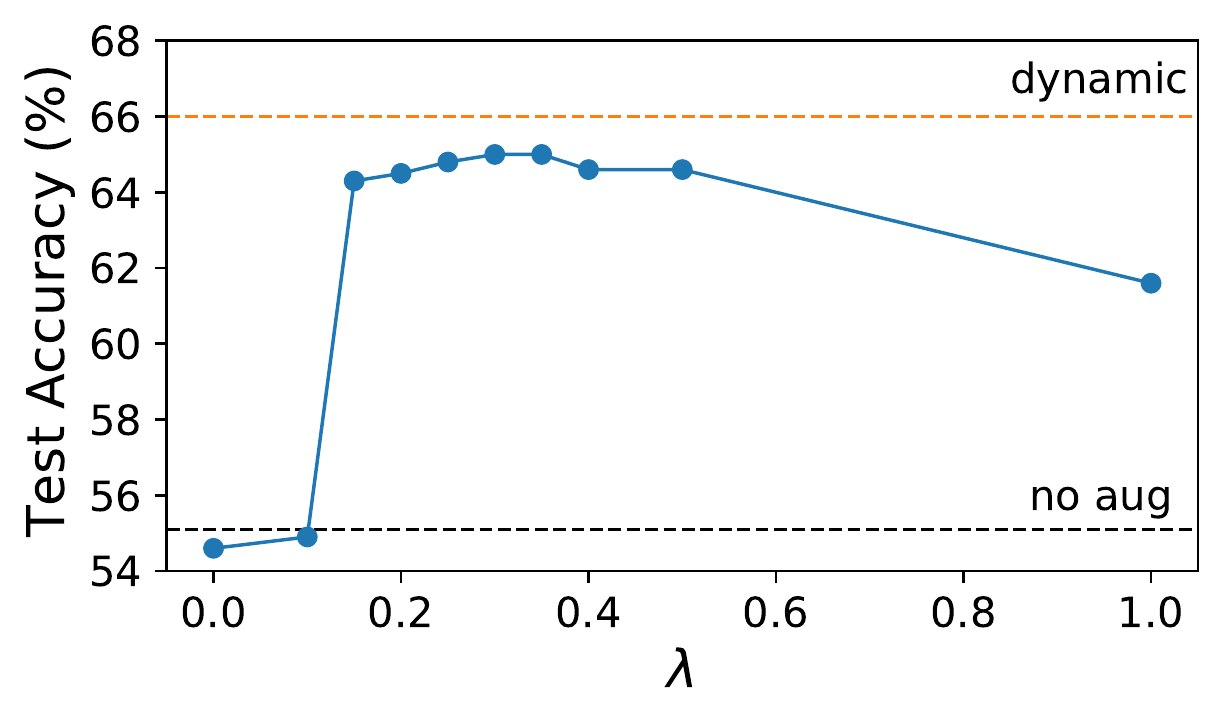}
 \caption{Model performance with fixed $\lambda$ on supervised cropping, compared with dynamic tuning $\lambda$ used in \method.
 }
 \label{fig:lambda_ablation}
\end{figure}

To show the effect of dynamically tuning $\lambda$ in \method, we compare it with results with fixed $\lambda$ in \cref{fig:lambda_ablation}. We find that for small $\lambda\le 0.1$, the entropy term is nearly 0 throughout training, which gives a result similar to no augmentation. For large $\lambda\ge 0.5$, the model suffers from excessive augmentations throughout training which hinders the training of the classifier and InstaAug module. There is also a performance plateau for $0.15\le\lambda\le 0.5$, whose accuracy is between $64.0$ and $65.0$. However, even the best of them is not as good as dynamic tuning, which is probably a result of their inability to keep transformation diversity stable during different stages of the training process.

\subsection{Why is the Random Augmentation baseline so strong?}
\label{app:sec:random}

It is perhaps initially surprising that the Random Augmentation baseline in~\ref{sec:exp_cropping} is so strong compared to the other global augmentation schemes.
In short, this occurs because the extensive hyperparameter sweep used for it turns out to be a more effective tuning mechanism than directly training global parameters simultaneously to the model.
To be more precise, for any \emph{global} cropping scheme (which includes random crop, Augerino, and InstaAug without input), there is little to be gained from using a non-uniform distribution on the position of the crops.  As such, the only thing that can be usefully learned is the distribution on the \emph{size} of the crops themselves. For the random crop baseline, we do an exhaustive sweep to establish the best distribution on crop sizes, meaning that this baseline represents a near-optimal global cropping augmentation.  By comparison, InstaAug (without input) must still learn the optimal cropping size distribution during training, and the results suggest that it does not always manage to do this perfectly, tending to prefer under-diverse transformations.  This is perhaps not surprising, as it does not have access to a validation set, unlike the hyperparameter sweep implicitly being deployed for the random crop baseline.  The problem is seen even more starkly for Augerino, where the lack of LRP causes training to become stuck in highly sub-optimal local optima that yield very little transformation diversity.

\end{appendices}

\end{document}